\pdfoutput=1

\documentclass[11pt]{article}

\usepackage[final]{latex/acl} %

\usepackage{times}
\usepackage{latexsym}

\usepackage[T1]{fontenc}

\usepackage[utf8]{inputenc}

\usepackage{microtype}

\usepackage{inconsolata}

\usepackage{amsmath}
\usepackage{amssymb}
\usepackage{booktabs}
\usepackage{enumitem}
\usepackage{graphicx}
\usepackage{listings}
\usepackage{soul}
\usepackage{tabularx}
\usepackage{csquotes}
\usepackage{hyperref}

\usepackage{tikz}
\usetikzlibrary{external}
\usetikzlibrary{shapes.geometric, arrows.meta, positioning, fit}
\usetikzlibrary{backgrounds}
\tikzset{
    token/.style={draw, rectangle},
    model/.style={draw, minimum width=2cm, minimum height=1cm},
    loss/.style={draw, circle},
}
\usepackage{pgfplots}
\usetikzlibrary{pgfplots.statistics}
\usepgfplotslibrary{colorbrewer}
\pgfplotsset{compat = 1.15, cycle list/Set1-8}
\usepackage{adjustbox}
\usetikzlibrary{plotmarks}
\usetikzlibrary{arrows, decorations.markings}
\usetikzlibrary{intersections}
\usepgfplotslibrary{fillbetween}
\usetikzlibrary{backgrounds}
\newcolumntype{Y}{>{\centering\arraybackslash}X}

\definecolor{custom_pink}{RGB}{250,108,115}
\definecolor{custom_purple}{RGB}{184,154,254}
\definecolor{custom_green}{RGB}{184,224,187}
\definecolor{custom_yellow}{RGB}{255,237,204}
\definecolor{custom_yellow2}{RGB}{255,218,146}
\definecolor{custom_turquoise}{RGB}{122,210,210}
\definecolor{custom_grey}{RGB}{234,232,232}
\definecolor{custom_dark_grey}{RGB}{138, 138, 138}

\definecolor{tree_fill}{RGB}{234,234, 234}
\definecolor{tree_draw}{RGB}{140, 137, 137}
\definecolor{tree_red}{RGB}{233, 157, 148}
\definecolor{tree_blue}{RGB}{161, 191, 225}
\definecolor{tree_red_dark}{RGB}{148,100,94}
\definecolor{tree_blue_dark}{RGB}{102,122,143}
\definecolor{tree_green_dark}{RGB}{108,132,83}

\definecolor{custom_palette_1}{RGB}{230, 57, 70}
\definecolor{custom_palette_2}{RGB}{241, 250, 238}
\definecolor{custom_palette_3}{RGB}{168, 218, 220}
\definecolor{custom_palette_4}{RGB}{69, 123, 157}
\definecolor{custom_palette_5}{RGB}{29, 53, 87}

\definecolor{custom_palette_3}{RGB}{168, 218, 220}

\definecolor{color_edge_list}{RGB}{254,196,114}
\definecolor{color_edge_list_dark}{RGB}{162,125,73}
\definecolor{color_goal}{RGB}{96,164,200}
\definecolor{color_path}{RGB}{170,208,131}

\definecolor{plot_grey}{RGB}{224,223,222}

\definecolor{grey}{RGB}{195,195,195}
\definecolor{red}{RGB}{233, 157, 148}
\definecolor{dark_red}{RGB}{148,100,94}
\definecolor{yellow}{RGB}{254,196,114}
\definecolor{yellow_dark}{RGB}{162,125,73}
\definecolor{green}{RGB}{170,208,131}
\definecolor{dark_green}{RGB}{108,132,83}
\definecolor{blue}{RGB}{161, 191, 225}
\definecolor{dark_blue}{RGB}{102,122,143}
\definecolor{pink}{RGB}{159,92,168}
\definecolor{orange}{RGB}{226,151,85}

\definecolor{tree_fill}{RGB}{234,234, 234}
\definecolor{tree_draw}{RGB}{140, 137, 137}

\newcommand{\token}[1]{
    \hspace{-0.1cm}\texttt{[#1}\texttt{]}\hspace{-0.1cm}
}
\newcommand{\subscripttoken}[2]{
    \hspace{-0.1cm}\texttt{[#1}$_{\text{\texttt{#2}}}$\texttt{]}\hspace{-0.1cm}
}

\makeatletter
\def\blfootnote{\xdef\@thefnmark{}\@footnotetext}
\makeatother

\title{A Mechanistic Analysis of a Transformer Trained on \\ a Symbolic Multi-Step Reasoning Task}

\author{Jannik Brinkmann$^{\dagger1}$ ~~~~~ Abhay Sheshadri$^{\dagger2}$ ~~~~~ Victor Levoso$^{\dagger3}$ \\
\textbf{Paul Swoboda}$^{4}$ ~~~~~ \textbf{Christian Bartelt}$^{1}$
\vspace{0.2cm} \\
\small{$^1$University of Mannheim ~~~ $^2$Georgia Institute of Technology ~~~ $^3$Independent\vspace{0.2cm}} ~~~ $^4$Heinrich-Heine University Düsseldorf  \\
}

\begin{document}
\maketitle
\begin{abstract}

\blfootnote{\hspace{-0.13cm}$^{\dagger}$\hspace{-0.086cm} Equal contribution. }
Transformers demonstrate impressive performance on a range of reasoning benchmarks.
To evaluate the degree to which these abilities are a result of actual reasoning, existing work has focused on developing sophisticated benchmarks for behavioral studies.
However, these studies do not provide insights into the internal mechanisms driving the observed capabilities.
To improve our understanding of the internal mechanisms of transformers, we present a comprehensive mechanistic analysis of a transformer trained on a synthetic reasoning task. 
We identify a set of interpretable mechanisms the model uses to solve the task, and validate our findings using correlational and causal evidence.
Our results suggest that it implements a depth-bounded recurrent mechanisms that operates in parallel and stores intermediate results in selected token positions.
We anticipate that the motifs we identified in our synthetic setting can provide valuable insights into the broader operating principles of transformers and thus provide a basis for understanding more complex models.\footnote{Correspondence to \href{mailto:jannik.brinkmann@uni-mannheim.de}{jannik.brinkmann@uni-mannheim.de}. The code is available at \href{https://github.com/abhay-sheshadri/backward-chaining-circuits}{github.com/backward-chaining-circuits}.}

\end{abstract}

\section{Introduction}
\label{cha:introduction}

Transformer-based language models~\citep{vaswani2017attention} demonstrate impressive performance on reasoning\footnote{In this paper, we consider a form of deductive reasoning as studied in \citet{saparov2023language}. For a discussion of different forms of reasoning, refer to \citet{reasoning_survey}.}\, tasks~\citep{kojima2023large}, mathematical problem-solving~\citep{cobbe2021training}, and planning~\citep{hung2022zeroshotplanner}.
However, despite strong performance on certain reasoning benchmarks, it remains unclear to what extent these abilities are a result of actual reasoning or simple pattern memorization~\cite{reasoning_survey}.

\begin{figure}[!t]
    \centering
    \hspace{0.12cm}\includegraphics{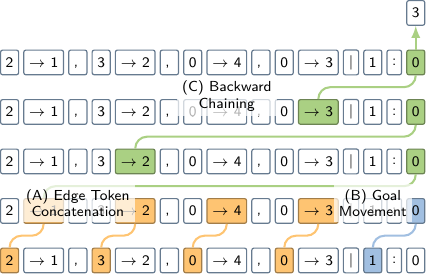}
    \vspace{0.1cm}
    \caption{Backward Chaining. 
    Given an input prompt, the model concatenates edge tokens in a single token position (A), and copies the goal node into the final token position (B). 
    The next step is then identified by applying an iterative algorithm that climbs the tree one level per layer (C).} 
    \label{fig:backward-chaining}
\end{figure}

To understand the reasoning capabilities of language models, existing work has focused on developing sophisticated benchmarks for behavioral studies~\citep{tafjord-etal-2021-proofwriter, saparov2023language, valmeekam2023planbench}.
However, the conclusions drawn from these studies \textit{do not provide insights into the internal mechanisms} driving the observed capabilities.
In contrast, recent work in the field of mechanistic interpretability attempts to understand the algorithms that models implement by reverse-engineering their internal mechanisms, and describing them at a certain level of abstraction~\citep{elhage2021mathematical}.
For example,~\citet{nanda2023progress} reverse-engineered how a small transformer model implements modular addition, providing insights into the specific computations performed by different components of the model. 
Similarly,~\citet{olsson2022incontext} discovered \enquote{induction heads} in transformers which enable a distinct copying mechanism that is considered to be crucial for the in-context learning abilities of language models. 

\paragraph{Contributions}
This paper studies reasoning in language models by reverse-engineering the internal mechanisms of a transformer trained on a symbolic multi-step reasoning task.
Specifically, we focus on path finding in a tree, a variation of the task proposed in \citet{saparov2023language}. 
By analyzing the internal representations of the model, we identify several key mechanisms: 
\begin{enumerate}
    \item A specific type of copying operation implemented in attention heads, which we call \textit{deduction heads}. 
    These are similar to induction heads as observed in~\citet{olsson2022incontext}. 
    In our task, deduction heads intuitively serve the purpose of moving one level up the tree. 
    These heads are implemented in multiple consecutive layers which allows the model to climb the tree multiple layers in a single forward pass. 
    \item A \textit{parallelization motif} whereby the early layers of the model choose to solve several subproblems in parallel that may be relevant for solving many harder instances of the task.
    \item A \textit{heuristic} for tracking the children of the current node and whether these children are leaf nodes of the tree.
    This mechanism is used when the model is unable to solve the problem using deduction heads and parallelization.
\end{enumerate}
We validate our findings using correlational and causal evidence, using techniques such as linear probing~\citep{alain2018understanding}, activation patching~\citep{vig-belinkov-2019-analyzing}, and causal scrubbing~\citep{chan_rigorously_nodate}.

\section{Related Work}
\label{cha:related_work}

\paragraph{Expressiveness of Transformers}
To understand the capabilities of transformers, one line of work characterizes their theoretical properties~\citep{bhattamishra-etal-2020-ability, merrill-etal-2021-provable, perez2021attention, merrill-etal-2022-saturated, liu2023transformers}.
These studies answer questions about the expressiveness of the transformer architecture by treating them as approximators of different classes of functions~\citep{Yun2020Are}, or by characterizing the class of formal languages that a transformer can recognize~\citep{strobl2023transformers}. 

\paragraph{Mechanistic Interpretability}
In contrast to these theoretical investigations, a number of studies have adopted an empirical approach in order to understand what transformers learn in practice~\cite{elhage2021mathematical, delétang2023neural, zhang2024transformerbased}. 
Our analysis is inspired by existing work in the field of mechanistic interpretability, attempting to discover and understand the algorithms implemented in a model by reverse-engineering its internal mechanisms~\citep{räuker2023transparent}. 
To explore these internal mechanisms, the field has adopted techniques such as activation patching~\cite{wang2023interpretability}, causal scrubbing~\citep{chan_rigorously_nodate}, and circuit discovery~\citep{conmy2023automated}.
In addition, considerable focus has been placed on the study of small models trained on specialized tasks, such as modular addition~\citep{nanda2023progress}, or group operations~\citep{chughtai2023neural}, providing a more manageable framework for understanding complex computational processes. 
We present an extended discussion of the related work on mechanistic interpretability in Appendix~\ref{app:mechanistic-interpretability}.

\paragraph{Evaluating Reasoning Capabilities}
Existing approaches to evaluate the reasoning capabilities of language models focus on their performance on a range of downstream tasks~\citep{reasoning_survey}.
To enable a more formal analysis of reasoning, a number of studies have developed sophisticated metrics and benchmarks~\citep{han2022folio, golovneva2023roscoe, wang-etal-2023-towards}.
For example,~\citet{saparov2023language} use a synthetic question-answering dataset based on a world model expressed in first-order logic to parse the generated reasoning processes into symbolic proofs for formal analysis.
Their results suggest that language models are capable of making correct individual deduction steps.
However, these approaches stop short of exploring the internal mechanisms that enable these capabilities~\citep{reasoning_survey}.
\citet{hou2023mechanistic} investigate whether language models solve reasoning tasks using information the model memorized from pretraining or using information provided in context.
To investigate this question, they use linear probes to determine whether the model encodes a reasoning tree.
The approach that comes closest to our work is~\citet{stolfo-etal-2023-mechanistic}, presenting a mechanistic interpretation of arithmetic reasoning by investigating the information flow in the model given simple mathematical questions.

\section{Background}
\label{cha:background}
\paragraph{Transformer Notation}
Transformers~\citep{vaswani2017attention} represent input text as a sequence~\( t_1, t_2, \ldots, t_N \) of tokens, such that \( t_i \in V \) where \( V \) is a vocabulary.
Each token $t_i$ is embedded as a vector \( \mathbf{x}^0_i \in \mathbb{R}^{d} \) using an embedding matrix~\( W_E \in \mathbb{R}^{|V| \times d} \), where $d$ is the dimension of the hidden state. 
These embeddings initialize the residual stream, which is then transformed through a sequence of $L$ transformer blocks, each consisting of a multi-head attention sublayer and an MLP sublayer.
The representation of token $t_i$ at layer $\ell$ is obtained by:
\begin{align}
    \mathbf{x}_i^\ell = \mathbf{x}_i^{\ell-1} + \mathbf{a}_i^\ell + \mathbf{m}_i^\ell
\end{align}
where $\mathbf{a}_i^\ell$ and $\mathbf{m}_i^\ell$ are the outputs of the attention and MLP sublayers.
To predict the next token in the sequence, it applies an unembedding matrix \( W_U \in \mathbb{R}^{|V| \times d} \) to the residual stream \( \mathbf{x}^L_i \), translating it into a probability distribution over the vocabulary.

\begin{figure*}[!t]
    \centering
    \includegraphics{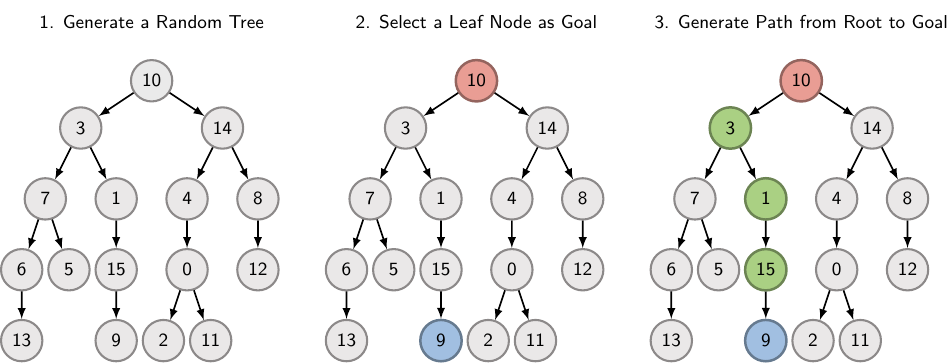}
    \vspace{-0.3cm}
    \caption{Data Generation. To generate our training set, we (1) generate a binary tree, (2) select a leaf node as the goal node, and (3) determine the path from the root to the goal node.}
    \label{fig:tree-generation}
    \vspace{0.4cm}
    \includegraphics{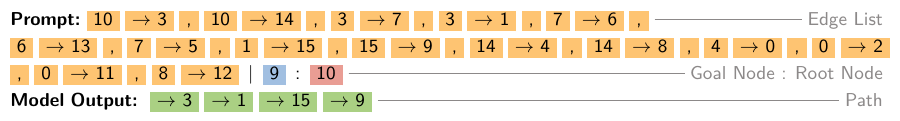}
    \vspace{-0.1cm}
    \caption{Prompt Format. The model receives input in a structured format, with each box representing a token. 
    The edge list of the tree is denoted as token pairs \subscripttoken{A}{1}\subscripttoken{B}{1}, $\dots$ \subscripttoken{A}{n}\subscripttoken{B}{n} followed by the task specification, including the goal \token{G} and the root node \subscripttoken{P}{1}.
    The model's objective is to predict the nodes in the path \subscripttoken{P}{2} $\dots$ \subscripttoken{P}{m1}, culminating in the goal node \subscripttoken{P}{m} = \token{G}.
    For simplification, our tokenization distinguishes tokens representing source and target nodes of each edge, such as \token{15} and \token{$\rightarrow$15}. }
    \label{fig:prompt-structure}
\end{figure*}

\paragraph{Linear Probes}
To investigate whether information is encoded in intermediate representations of the model, we use linear probes~\citep{alain2018understanding} implemented as a linear projection from the residual stream.
This involves training a logistic regression model on a dataset of activations \( \mathbf{x}_i^\ell\ \) to predict an auxiliary classification problem.
For example, we use it to determine whether information about an edge is encoded in the activations at a specific token position. 
If the performance of the linear probe is sufficiently high, we can treat this as evidence for the information being encoded.

\paragraph{Activation Patching}
To evaluate the importance of a model component for a given prediction, we intervene by patching in the activations it would have had on a different input~(also called resampling ablations)~\citep{NEURIPS2020_92650b2e, meng2022locating}.
This involves using a clean input~$s$ with an associated target prediction~$r$, and a corrupted input~$s'$ with a different target~$r'$. 
Then, we cache the activations of the component on~$s$, and evaluate the effect of patching in these activations when running the model on~$s'$.
To compute the effect of this intervention, we compute the difference in logits:
\begin{align}
    LD(r, r') = \text{Logit}(r) -\text{Logit}(r')
\end{align}

\paragraph{Causal Scrubbing}
To evaluate specific hypotheses about internal mechanisms, we use causal scrubbing which evaluates the effect of behavior-preserving resampling ablations~\citep{chan_rigorously_nodate}. 
Specifically, given a hypothesis about which component of a model implements a specific behavior, we replace the activations of that component on some input with activations on another input, where our hypothesis predicts that the activations represent the same thing.
Then, we evaluate the impact of this intervention by computing how much of the initial performance is recovered: 
\begin{align}
    L_{CS} = \frac{L_{scrubbed} - L_{random}}{L_{model} - L_{random}}
\end{align}
where $L_{model}$ is the test loss of the trained model, $L_{random}$ of a model that outputs uniform logits, and $L_{scrubbed}$ of the trained model with the chosen activations resampled.
In contrast to activation patching, which provides insights about whether a specific activation is causally linked to the output, causal scrubbing provides stronger evidence about the role of the activations in the model. 
Recovering the majority of the loss with a given hypothesis demonstrates that the activation corresponds to a specific variable in a high-level causal model.

\section{Experimental Setup}
\label{cha:experimental_setup}
\subsection{Task Description}
\label{cha:task-description}
We focus on path finding in trees as a modified version of the task proposed by~\citet{saparov2023language}.
Our adaptation shifts the focus from reasoning in natural language to abstract symbolic reasoning. 
This allows us to better understand motifs that the models might be using to solve analogous problems in natural language. 
In our experimental setup, we generate training samples by generating binary trees \( T = (V, E) \) uniformly at random from the set of all trees with 16 nodes, i.e.\ \(|V| = 16\).
Then, for each tree, a leaf node is randomly selected as the target node (see Figure~\ref{fig:tree-generation}).
The model is given the edge list \subscripttoken{A}{1}\subscripttoken{B}{1}, \subscripttoken{A}{2}\subscripttoken{B}{2}, $\dots$ \subscripttoken{A}{n}\subscripttoken{B}{n} of the tree, a specified goal \token{G}, and the root node \subscripttoken{P}{1}, and is trained to predict the path from the root node to the goal \subscripttoken{P}{2}\subscripttoken{P}{3} $\dots$ \subscripttoken{P}{m} such that \subscripttoken{P}{m} $=$ \token{G}, as depicted in Figure~\ref{fig:prompt-structure}.
The path between any two nodes in a tree is unique; therefore, the data generation is not affected by the type of search algorithm used to determine the path.
We emphasize that this task is nontrivial, as the model has to perform multiple reasoning steps in a single forward pass to predict the next step.
At each step, the model must determine from which node the goal is reachable.
Thus, the model must perform multiple steps of reasoning for each next token prediction \textit{without} relying on techniques such as chain-of-thought or scratchpad~\citep{wei2022chain}, which would allow the model to roll-out the reasoning steps over multiple next token predictions.

\subsection{Model Specification and Training Process}
\label{cha:model-training}
In our experiments, we use a 6-layer, decoder-only transformer with an embedding dimension of 128, a single attention head per layer, and a feed-forward dimension of 512, resulting in a total of 1.2 million parameters.
The training dataset consists of~150,000 generated trees.
The edge lists of these trees are shuffled to prevent the model from learning simple heuristics and encourage structural understanding of trees.
To evaluate the performance of our model, we compute the accuracy based on the exact match of complete sequences using greedy decoding.
Our model achieves 99.7~\hspace{-0.05em}\% accuracy rate on a test set of 15,000 unseen trees, despite seeing just a small fraction of all possible trees during training (see Appendix~\ref{app:experimental-setup}).
This suggests that generalization is required for meaningful performance and that the model has learned to be capable of solving pathfinding in trees.

\section{Symbolic Reasoning using Backward Chaining}
\label{cha:findings}
In this section, we present a mechanistic analysis of the internal mechanisms of the model and provide correlational and causal evidence.
Our findings suggest that the model uses an interpretable and meaningful backward chaining algorithm to perform pathfinding in a tree. 
To help guide the reader, we present an intuitive explanation before breaking down the individual algorithmic steps in the following sections. 

\paragraph{The Backward Chaining Algorithm}
First, the model aggregates the source and target nodes of each edge in the edge list into the target node position (see Section~\ref{cha:edge-token-concatenation}). 
Then, the model starts at the goal node and moves up the tree one level with each layer of the model. 
This mechanism is implemented using attention heads, which we term \enquote{deduction heads}~(see Section~\ref{cha:backward-chaining}). 
By the composition of multiple attention heads in consecutive layers, the model can traverse the tree upwards for~$L-1$ edges, where $L$ is the number of layers in the model.
We refer to this mechanism as backward chaining, inspired by the use of the term in the symbolic artificial intelligence literature~\citep{Russell2009-ht}.
In more complex scenarios, where the required path exceeds the depth of the model, it relies on backward chaining from multiple nodes in the tree in parallel. 
This creates multiple subpaths that can be used to find the correct next step in cases where the goal node is more than~$L-1$ edges away (see Section~\ref{cha:path-merging}).
In addition, the model uses a simple heuristic as a fallback mechanism, where it identifies child nodes of the current position and evaluates whether these are leaf nodes of the tree.
This enables the model to make informed guesses when backward chaining and parallelization are insufficient to solve the problem (see Section~\ref{cha:one-step-lookahead}).

\subsection{Edge Token Concatenation}
\label{cha:edge-token-concatenation}
The attention head in the first layer of the model creates edge embeddings by moving the information about the source token onto the target token for each edge in the context.
Specifically, for each edge \token{A}\token{B} it copies the information from \token{A} into the residual stream at position \token{B}.
This mechanism has some similarities with ``Previous Token Heads'', as observed in pre-trained language models~\citep{olsson2022incontext, wang-etal-2023-towards}. 

\paragraph{Experiment: Linear Probes}
To validate that the model creates edge embeddings, we train a linear probe to predict the associated edge given the activations $\mathbf{x}^1$ at the positions of target nodes. 
The probe is trained using 8,000 examples and evaluated on a test dataset of the same size. 
For comparison, we also report the performance of a linear probe given the activations $\mathbf{x}^0$ at the positions of the target nodes and probes given the activations at the positions of the source node.

\paragraph{Results}
Table~\ref{tab:linear-probe-edge-embeddings} reports the performance of the linear probe measured using the F1 score.
We find that we can successfully extract the source and target tokens \token{A}\token{B} from the residual stream activations $\mathbf{x}^1$ at the position of the target tokens after the first layer, providing strong evidence for the edge token concatenation hypothesis as described above. 
Moreover, it does not encode the complete edge in the position of the source token, attributed to causal masking in the attention mechanism.

\begin{figure*}[!t]
    \centering
    \includegraphics{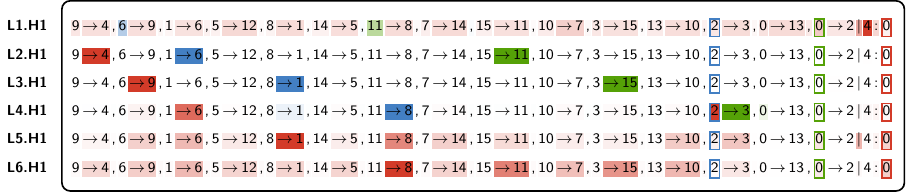}
    \caption{Visualization of multi-layer attention patterns on an example input. We show the attention from three selected positions: the \textcolor{red!200}{path position}, \textcolor{blue!200}{register token at position 39}, and \textcolor{green!200}{register token at position 44}. We show that the path node starts backward chaining from the specified goal, while the two register tokens start backward chaining from different subgoals. Each token is highlighted by the color of the token that most strongly attends to it.  The intensity of the color is based on the magnitude of the attention score. For details on how we select the register tokens and more examples, see Appendix~\ref{app:attention-patterns}.}
    \label{fig:attention-patterns}
\end{figure*}

\subsection{Backward Chaining}
\label{cha:backward-chaining}
The most important mechanism the model uses to predict the correct next step is an iterative algorithm, which we refer to as backward chaining.
The algorithm starts at the goal node and climbs the tree one level per layer.
To this end, the model copies the target node \token{G} into the final token position \subscripttoken{P}{i} (see Table~\ref{tab:linear-probe-edge-embeddings}) and then in each consecutive layer applies what we term \enquote{deduction heads}.

\paragraph{Mechanism: Deduction Heads}
The function of deduction heads is to search for the edge in the context in which the current position is the target node \token{B}, find the corresponding source token \token{A}, and then copy the source token over to the current position.
Thus, deduction heads complete the pattern by mapping:
\begin{align}
   \token{A}\enspace\; \token{B}\enspace ...\enspace \token{B}\enspace \rightarrow\enspace \token{A}
\end{align}
In other words, this mechanism enables the model to go one step up the tree and append \token{A} after having seen the last \token{B} in the sequence.
This mechanism depends on the edge-token concatenation, which previously copied information about \token{A} into \token{B} (see Appendix~\ref{app:head-composition}). 
This allows the deduction head to search for information about \token{B} but then copy information about \token{A} into the final token position. 

By composition of multiple deduction heads in consecutive layers, the model is capable of climbing several steps up the tree in a single inference step. 
This creates a subpath at the final token position whose lengths is equivalent to the number of layers involved.
In our model, we observe that the attention heads after the first layers can act as deduction heads, resulting in a backward chaining depth of at most $L - 1$ steps. 
The role of these heads depends on the proximity of the current position to the goal node; for example, if the root node is just four steps away from the goal node, the model recognizes this and the attention head in the sixth layer does not act as a deduction head.

\begin{table}[!t]
    \centering
    \small
    \vspace{0.1cm}
    \caption{F1 score of linear probes trained to predict the edge \token{A}\token{B} given the residual stream activations at position \token{A} or \token{B}. In addition, we report the performance of a linear probe to predict the goal node \token{G} from the positions in the path \subscripttoken{P}{i}.}
    \begin{tabular}{p{0.46\columnwidth} >{\centering\arraybackslash}p{0.12\columnwidth} >{\centering\arraybackslash}p{0.12\columnwidth}}
        \toprule
         & $\mathbf{x}_i^0$ & $\mathbf{x}_i^1$ \\
         \midrule
         Linear \{\subscripttoken{A}{i} $\rightarrow$ \subscripttoken{A}{i}\subscripttoken{B}{i}\} & \textsc{0.13} & \textsc{0.19} \\
         Linear \{\subscripttoken{B}{i} $\rightarrow$ \subscripttoken{A}{i}\subscripttoken{B}{i}\} & \textsc{0.11} & \textbf{\textsc{1.00}} \\
         \midrule
         Linear \{\subscripttoken{P}{i} $\rightarrow$ \token{G}\} & \textsc{0.03} & \textbf{\textsc{1.00}} \\
         \bottomrule
    \end{tabular}
    \vspace{-0.1cm}
    \label{tab:linear-probe-edge-embeddings}
\end{table}

\paragraph{Experiment: Causal Scrubbing}
To confirm that the model uses backward chaining to predict the next step for paths up to a depth of $L-1$, we use causal scrubbing~(see Section~\ref{cha:background}).
Specifically, we hypothesize that the attention head of layer $\ell$ is responsible for writing the node that is $\ell - 1$ edges above the goal into the final token position in the residual stream.
This implies that the output of the attention head in layer $\ell$ should be consistent across trees that share the same node $\ell - 1$ edges above the goal.
To test this, we generate a clean and a corrupted graph that share the same node $\ell - 1$ edges above the goal node. 
Then, we substitute the output of the head on the clean graph with the output of the head on the corrupted graph, and measure the difference in the loss.

\begin{figure}[!ht]
    \centering
    \vspace{0.2cm}
    \includegraphics{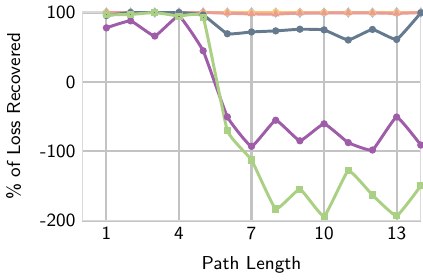}
    \caption{To test whether the model predicts the next step using backward chaining, we perform resampling ablations on each head using causal scrubbing. 
    We find that we can recover close to 100 \% of the performance of the model for paths up to length $L-1$, providing strong evidence for our backward chaining hypothesis.}
    \label{fig:backward-chaining-causal-scrubbing}
    \vspace{-0.2cm}
\end{figure}

\paragraph{Results}
Figure~\ref{fig:backward-chaining-causal-scrubbing} illustrates the effect of causal scrubbing. 
We find that we can recover most of the performance (close to 100 \%) of the model for paths up to length $L-1$, providing strong evidence for our hypothesis about backward chaining. 
We also find that this hypothesis explains most of the behavior of the attention heads in the first four layers of the model even on paths that require more than $L-1$ steps; only the attention heads in the final two layers act significantly different, such that the model ends up confidently making incorrect predictions after we apply causal scrubbing.
This highlights that the heads in these two layers cannot be accurately described as deduction heads over the entire data distribution.

\subsection{Path Merging}
\label{cha:path-merging}
In cases where the goal is more than $L-1$ steps away from the current position, the previously described mechanism is insufficient. 
To address this, the model does not only perform backward chaining on the final token position, but in parallel on multiple different token positions.
We refer to these tokens as \enquote{register tokens}.
The resulting subpaths are then merged on the final token position to facilitate more complex scenarios.

\paragraph{Observation: Register Tokens}
The role of register tokens is to act as working memory.
These are tokens that do not contain any useful information for the actual task; either they do not contain any useful information to begin with, e.g.\ \token{,}, or are tokens whose information has been copied to other positions and thus contain redundant information.
For more details on the role of register tokens in our context, see Appendix \ref{app:register-tokens}.

\paragraph{Mechanism: Path Merging}
To compute paths for which backward chaining on the final token position is not sufficient, the model uses register tokens to perform backward chaining in parallel from multiple positions in the tree. 
To this end, similar to backward chaining from the goal, it copies a node into each of these register tokens which will then be picked up by the deduction heads. 
This results in a multiple subpaths being stored at different positions in the sequence. 
Then, the model can merge these by finding overlapping subpaths.

To illustrate, let us assume that a subpath \token{B}$\rightarrow$\token{C}$\rightarrow$\token{G} has been stored in the final token position and a different subpath \token{A}$\rightarrow$\token{E}$\rightarrow$\token{B} has been stored in some register token.
Then, during path merging, the model copies the subpath stored in the register token into the final token position, enabling it to move up the tree multiple steps at a time. 

\paragraph{Experiment: Linear Probe}
To validate that the model is indeed generating subpaths in register tokens, we train a linear probe to predict these subpaths given the residual stream at these register token positions. 
We generate the training set for the linear probe by inspecting the attention patterns~(see Figure~\ref{fig:attention-patterns}).
We observe that these approximate hard-attention, which allows us to read off the token positions from which the head is copying information.
We extract the subpath we would expect to be encoded at each of these register tokens and transform them into an adjacency matrix representation. 
Then, we train the probe on a set of~8,000~examples and evaluate it on a test set of the same size.
The probe achieves an F1 score of~92.82~\% when trained on the activations of layer four, i.e. up to the the point where the model is performing backward chaining.

\begin{table}[!t]
    \centering
    \small
    \caption{F1 score of a linear probe trained to predict the adjacency matrix of the subpath we hypothesise to be encoded in the activations $x_i^4$ at register token positions.}
    \begin{tabular}{p{0.46\columnwidth} >{\centering\arraybackslash}p{0.12\columnwidth} >{\centering\arraybackslash}p{0.12\columnwidth}}
        \toprule
         & $\mathbf{x}_{i}^{4}$ \\
         \midrule
         Linear \{\subscripttoken{R}{i} $\rightarrow$ \subscripttoken{S}{i}\} & \textbf{\textsc{92.82}} \\
         \bottomrule
    \end{tabular}
    \vspace{-0.1cm}
    \label{tab:linear-probe-register-tokens}
\end{table}

\paragraph{Experiment: Register Token Patching}
To evaluate whether the subpaths stored in the register tokens have a causal effect on the prediction of the model, we perform resampling ablations (see Chapter~\ref{cha:background}) on the register tokens positions in trees which
(i)~contain sufficiently long paths such that backward chaining on the final token position is insufficient, and 
(ii)~positions where nodes have multiple child nodes, ensuring that the model has to make a decision between multiple options. 
The corrupted activations are extracted from another tree in the same class as described above.
We compute the effect of this intervention using the logit difference.  
We perform this intervention after layer four, as we found that the behavior of all previous layers is explained using our backward chaining hypothesis (see Section~\ref{cha:backward-chaining}).

\begin{figure}[!ht]
    \centering
    \vspace{0.2cm}
    \includegraphics{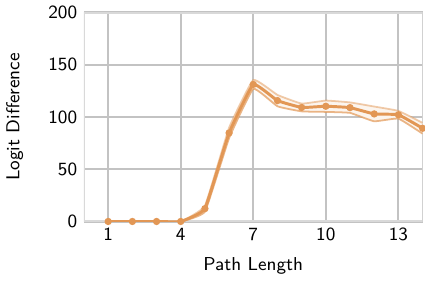}
    \caption{
    To test whether the model relies on subpaths stored in register tokens, we perform resampling ablations on the register token positions at $\mathbf{x}_i^4$.
    Here, we conducted 10 separate runs, each involving 1000 samples. 
    For each run, we calculate the mean logit difference and report the 95 \% confidence interval for the average effects observed across the runs.
    The results demonstrate that these subpaths are instrumental for paths longer than $L - 1$ steps.
    }
    \label{fig:path-merging-activation-patching}
    \vspace{-0.2cm}
\end{figure}

\paragraph{Results}
Figure~\ref{fig:path-merging-activation-patching} illustrates the impact of patching the register tokens on the model predictions at different path lengths. 
Our results show that the intervention has no effect on performance up to a path depth of 4 and minimal effect at depth 5, which is consistent with our backward chaining hypothesis. 
Beyond this depth, this intervention has a significant effect on the performance.
This suggests that the encoded subpaths are causally relevant for predicting next steps on paths that are more than $L-1$ steps away from the goal. 
However, our findings also indicate that the predictions are not solely dependent on these subpaths derived, but other factors besides the subpaths contribute to the prediction.
This includes the influence of a one-step lookahead mechanism, which identifies child nodes of the current position and increases the probabilities of the children that are not leaves.

\subsection{One-Step Lookahead}
\label{cha:one-step-lookahead}
We find that the model uses an additional mechanism, which identifies child nodes of the current position and increases the prediction probabilities of the children that are not leaf nodes of the tree.
This enables the model to make informed guesses in cases where backward chaining is not sufficient.
This mechanism is particularly effective on long paths as these have a lower branching factor in our experimental setup.
Thus, it is a pragmatic strategy to minimize the training error. 

\begin{figure*}[!t]
    \centering
    \includegraphics[width=0.8\textwidth]{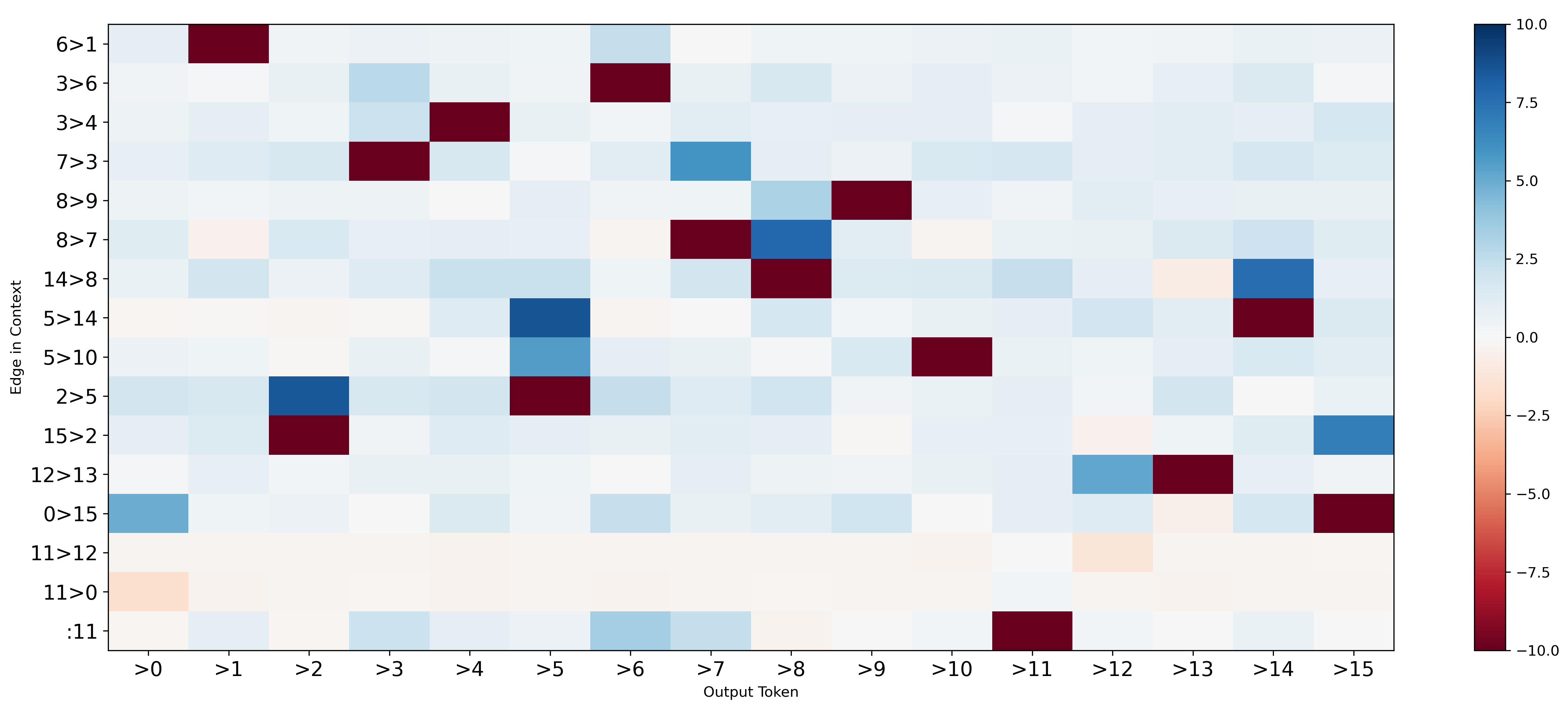}
    \caption{Contribution of each target node position to the logits at the path position through the attention heads \texttt{L5.H1} and \texttt{L6.H1} on a specific example.}
    \label{fig:dla-backup}
\end{figure*}

\paragraph{Experiment: Linear Probes}
To validate that the model represents the child nodes of the current position, including whether they are leaf nodes, we use linear probes. 
The probes are trained to predict this information given the activations on the final token position. 
Table~\ref{tab:linear-probe-lookahead} reports the performance of the linear probes measured using the F1 score. 
Our analysis shows that the model starts to collect information about the children and leaf nodes from the fourth layer and represents both aspects in the fifth layer.

\begin{table}[!ht]
    \centering
    \small
    \vspace{0.1cm}
    \caption{F1 score of linear probes trained to predict the children of the current position and whether these are leafs of the tree given the residual stream activations at position \subscripttoken{P}{i}.}
    \begin{tabular}{l >{\centering\arraybackslash} c >{\centering\arraybackslash} c >{\centering\arraybackslash} c}
        \toprule
         & $\mathbf{x}_{i}^{4}$ & $\mathbf{x}_{i}^{5}$ & $\mathbf{x}_{i}^{6}$ \\
         \midrule
         Linear \{\subscripttoken{P}{i} $\rightarrow$ \subscripttoken{Children}{i}\} & \textsc{0.00} & \textsc{47.88} & \textbf{\textsc{98.20}} \\
         Linear \{\subscripttoken{P}{i} $\rightarrow$ \subscripttoken{Leafs}{i}\} & \textsc{0.00} & \textsc{49.71} & \textbf{\textsc{95.76}} \\
         \bottomrule
    \end{tabular}
    \vspace{-0.1cm}
    \label{tab:linear-probe-lookahead}
\end{table}

\paragraph{Mechanism}
The results from linear probes suggest that the attention heads in the final two layers of the model are responsible for the one-step lookahead mechanism. 
To better understand this mechanism, we examine how these two attention heads directly compose with the unembedding matrix.
Specifically, we compute the contribution of each token to the logits through these attention heads.
Since each edge will be represented in the target node (see Section~\ref{cha:edge-token-concatenation}), we focus on target node positions (see Figure \ref{fig:dla-backup}).
By inspecting their query-key (QK) circuits~\cite{elhage2021mathematical}, we find that these two heads attend to the target node of every edge except those for which the source node is the current path position. 
Furthermore, we can break the mechanism in their output-value (OV) circuits into three components: 
\begin{enumerate}[leftmargin=*]
    \itemsep0em
    \item Each edge decreases the logit of its target node.
    \item Each edge increases the logit of its source node.
    \item Each token in the path decreases its logit.
\end{enumerate}
As a result of these components, the logits of the leaf nodes in the tree will decrease while the logits of the children of the current position will increase. 
For the other nodes, the logit increase from being a parent, and the logit decrease from being a child node which roughly cancel out, causing their logit to remain constant.

\section{Discussion}
\label{cha:discussion}
\paragraph{Register Tokens}
Our model uses some token positions as a form of working memory to store intermediate results.
This observation aligns with \citet{darcet2023vision} which found that image models use some image patches to accumulate global information while discarding spatial information. 
Similarly, \citet{goyal2023think} show that adding uninformative tokens at the end of each prompt can enhance language model performance on downstream tasks without introducing additional parameters. 
Our findings suggest that these techniques enable the model to store intermediate results and perform more computations in parallel.
This is consistent with theoretical insights from \citet{merrill-etal-2022-saturated} which highlights how the effective state of a transformer depends on the number of tokens in the sequence.

\paragraph{Structural Recursion}
Transformers, which are by definition non-recurrent, struggle with emulating structural recursion and extracting recursive rules from data~\citep{zhang2024transformerbased}.
This aspect of learning is crucial in domains such as programming and formal mathematics, where understanding complex relationships relies on these abilities.
Our analysis provides insights into possible reasons for this limitation. 
In our setting, training the model using standard objectives for next-token prediction forces the model to unroll the entire recursive structure in a single forward pass.
This restricts their abilities to process recursion, leading them to resort to shortcut solutions~\citep{liu2023transformers}. 

\paragraph{Reasoning in Transformers}
There is an ongoing debate about the reasoning capabilities of transformers~\cite{reasoning_survey}.
Some argue that these models might just be capable of memorizing patterns without gaining causal understanding, which could lead to diminishing performance on out-of-distribution data~\cite{bender-koller-2020-climbing, Floridi2020-FLOGIN, bender2021, merrill-etal-2021-provable}. 
However, there are several observations that suggest that transformers might be capable of more than just pattern recognition; e.g. \citet{olsson2022incontext} found a simple algorithm implemented in attention heads that contributes to the in-context learning abilities of transformers and operates independent of the specific tokens. 
This algorithm is doing more than memorizing patterns and can in some sense work out-of-distribution. 
In our synthetic setting, we found that the model learned an interpretable and meaningful backward chaining algorithm, supporting the claim that transformers might be capable of a form of reasoning that goes beyond simple pattern memorization.
These results further complement the findings of \citet{hou2023mechanistic}, shedding light on the mechanisms transformer models might use to compute reasoning trees over information provided in context.
However, it is important to note that findings from our synthetic settings do not support the boarder claim that transformers possess general reasoning capabilities, highlighting the need for further investigations.

\section{Conclusion}
\label{cha:conclusion}

In this paper, we conducted a mechanistic analysis of a transformer trained on pathfinding in trees.
Our results suggest that the model implements a backward chaining algorithm that starts at the goal node and climbs the tree one level per layer. 
To solve more complex problems, where the required reasoning depth exceeds the number of layers, it executes the same mechanism in parallel across multiple register tokens and combines the resulting subpaths on the final token position. 
In addition, it performs a simple one-step lookahead in which it finds the child nodes of the current position and evaluates whether they are leaf nodes. 

Our findings in this synthetic setting demonstrate the ability of a transformer to perform deductive reasoning up to a certain reasoning depth, after which it resorts to simple heuristics.
By using parallelized computations to store intermediate results in register tokens and then merging these results in the final token position, the model demonstrates a form of deductive reasoning that, while effective within a given setting, is constrained by its architecture.
These observations suggests that transformers may exhibit a inductive bias towards adopting highly parallelized strategies for tasks involving search, planning, or reasoning.

\section*{Limitations}
\label{cha:limitations}
\paragraph{Synthetic Task}
Our experiments were conducted on a symbolic reasoning task (see Chapter \ref{cha:task-description}).
This allowed us to bypass the complexities associated with natural language, such as multi-token embeddings~\citep{nanda2023factfinding}.
In addition, our tokenization distinguishes tokens representing source and target nodes of each edge, such as \token{15} and \token{$\rightarrow$15}. 
Therefore, our findings are specific to our model and it remains unclear whether large language models trained on natural language use similar mechanisms to solve this task. 
However, we anticipate that the motifs we discovered in our synthetic setting can provide valuable insights into the broader operating principles of transformers and thus provide a basis for understanding more complex models.

\paragraph{Input Format}
To prevent the model from learning shortcuts based on the order of the edges in the prompt, we trained our model on shuffled edge lists~(see Section~\ref{cha:model-training}). 
However, our analysis is limited to sequences in which the edge list is presented in backward order. 
By backward order we mean a listing of edges that starts with the leaf nodes and ascends level by level to the root node, as opposed to a forward order where the listing starts with the root node and progresses downwards through each level. 
Our investigation does not extend to a detailed examination of alternative arrangements of the edge list.
However, preliminary observations suggest that the model uses similar mechanisms with minor variations, such as the use of different register tokens.

\section*{Acknowledgements}
Jannik Brinkmann is supported by the German Federal Ministry for Digital and Transport (BMDV) and the German Federal Ministry for Economic Affairs and Climate Action (BMWK). 
Abhay Sheshadri and Victor Levoso have been supported by Lightspeed Grants.

\bibliography{custom}

\begin{thebibliography}{57}
\expandafter\ifx\csname natexlab\endcsname\relax\def\natexlab#1{#1}\fi

\bibitem[{Alain and Bengio(2018)}]{alain2018understanding}
Guillaume Alain and Yoshua Bengio. 2018.
\newblock \href {http://arxiv.org/abs/1610.01644} {Understanding intermediate layers using linear classifier probes}.

\bibitem[{Belrose et~al.(2023)Belrose, Furman, Smith, Halawi, Ostrovsky, McKinney, Biderman, and Steinhardt}]{belrose2023eliciting}
Nora Belrose, Zach Furman, Logan Smith, Danny Halawi, Igor Ostrovsky, Lev McKinney, Stella Biderman, and Jacob Steinhardt. 2023.
\newblock \href {http://arxiv.org/abs/2303.08112} {Eliciting latent predictions from transformers with the tuned lens}.

\bibitem[{Bender et~al.(2021)Bender, Gebru, McMillan-Major, and Shmitchell}]{bender2021}
Emily~M. Bender, Timnit Gebru, Angelina McMillan-Major, and Shmargaret Shmitchell. 2021.
\newblock \href {https://doi.org/10.1145/3442188.3445922} {On the dangers of stochastic parrots: Can language models be too big?}
\newblock In \emph{Proceedings of the 2021 ACM Conference on Fairness, Accountability, and Transparency}, FAccT '21, page 610–623, New York, NY, USA. Association for Computing Machinery.

\bibitem[{Bender and Koller(2020)}]{bender-koller-2020-climbing}
Emily~M. Bender and Alexander Koller. 2020.
\newblock \href {https://doi.org/10.18653/v1/2020.acl-main.463} {Climbing towards {NLU}: {On} meaning, form, and understanding in the age of data}.
\newblock In \emph{Proceedings of the 58th Annual Meeting of the Association for Computational Linguistics}, pages 5185--5198, Online. Association for Computational Linguistics.

\bibitem[{Bhattamishra et~al.(2020)Bhattamishra, Ahuja, and Goyal}]{bhattamishra-etal-2020-ability}
Satwik Bhattamishra, Kabir Ahuja, and Navin Goyal. 2020.
\newblock \href {https://doi.org/10.18653/v1/2020.emnlp-main.576} {On the {A}bility and {L}imitations of {T}ransformers to {R}ecognize {F}ormal {L}anguages}.
\newblock In \emph{Proceedings of the 2020 Conference on Empirical Methods in Natural Language Processing (EMNLP)}, pages 7096--7116, Online. Association for Computational Linguistics.

\bibitem[{Catalan(1838)}]{Catalan1838}
Eug{\`e}ne~Charles Catalan. 1838.
\newblock Note sur une {\'e}quation aux diff{\'e}rences finies.
\newblock \emph{Journal de Math{\'e}matiques Pures et Appliqu{\'e}es}, 3:508--516.

\bibitem[{Chan et~al.(2022)Chan, Garriga-alonso, Goldowsky-Dill, Greenblatt, Jenny, Radhakrishnan, Buck, and Thomas}]{chan_rigorously_nodate}
Lawrence Chan, Adrià Garriga-alonso, Nicholas Goldowsky-Dill, Ryan Greenblatt, Jenny, Ansh Radhakrishnan, Buck, and Nate Thomas. 2022.
\newblock \href {https://www.lesswrong.com/posts/JvZhhzycHu2Yd57RN/causal-scrubbing-a-method-for-rigorously-testing} {Causal {Scrubbing}: a method for rigorously testing interpretability hypotheses}.

\bibitem[{Chughtai et~al.(2023)Chughtai, Chan, and Nanda}]{chughtai2023neural}
Bilal Chughtai, Lawrence Chan, and Neel Nanda. 2023.
\newblock \href {https://openreview.net/forum?id=j4_YHiTAN63} {Neural networks learn representation theory: Reverse engineering how networks perform group operations}.
\newblock In \emph{ICLR 2023 Workshop on Physics for Machine Learning}.

\bibitem[{Cobbe et~al.(2021)Cobbe, Kosaraju, Bavarian, Chen, Jun, Kaiser, Plappert, Tworek, Hilton, Nakano, Hesse, and Schulman}]{cobbe2021training}
Karl Cobbe, Vineet Kosaraju, Mohammad Bavarian, Mark Chen, Heewoo Jun, Lukasz Kaiser, Matthias Plappert, Jerry Tworek, Jacob Hilton, Reiichiro Nakano, Christopher Hesse, and John Schulman. 2021.
\newblock \href {http://arxiv.org/abs/2110.14168} {Training verifiers to solve math word problems}.

\bibitem[{Conmy et~al.(2023)Conmy, Mavor-Parker, Lynch, Heimersheim, and Garriga-Alonso}]{conmy2023automated}
Arthur Conmy, Augustine~N. Mavor-Parker, Aengus Lynch, Stefan Heimersheim, and Adrià Garriga-Alonso. 2023.
\newblock \href {http://arxiv.org/abs/2304.14997} {Towards automated circuit discovery for mechanistic interpretability}.

\bibitem[{Darcet et~al.(2023)Darcet, Oquab, Mairal, and Bojanowski}]{darcet2023vision}
Timothée Darcet, Maxime Oquab, Julien Mairal, and Piotr Bojanowski. 2023.
\newblock \href {http://arxiv.org/abs/2309.16588} {Vision transformers need registers}.

\bibitem[{Delétang et~al.(2023)Delétang, Ruoss, Grau-Moya, Genewein, Wenliang, Catt, Cundy, Hutter, Legg, Veness, and Ortega}]{delétang2023neural}
Grégoire Delétang, Anian Ruoss, Jordi Grau-Moya, Tim Genewein, Li~Kevin Wenliang, Elliot Catt, Chris Cundy, Marcus Hutter, Shane Legg, Joel Veness, and Pedro~A. Ortega. 2023.
\newblock \href {http://arxiv.org/abs/2207.02098} {Neural networks and the chomsky hierarchy}.

\bibitem[{Elhage et~al.(2021)Elhage, Nanda, Olsson, Henighan, Joseph, Mann, Askell, Bai, Chen, Conerly, DasSarma, Drain, Ganguli, Hatfield-Dodds, Hernandez, Jones, Kernion, Lovitt, Ndousse, Amodei, Brown, Clark, Kaplan, McCandlish, and Olah}]{elhage2021mathematical}
Nelson Elhage, Neel Nanda, Catherine Olsson, Tom Henighan, Nicholas Joseph, Ben Mann, Amanda Askell, Yuntao Bai, Anna Chen, Tom Conerly, Nova DasSarma, Dawn Drain, Deep Ganguli, Zac Hatfield-Dodds, Danny Hernandez, Andy Jones, Jackson Kernion, Liane Lovitt, Kamal Ndousse, Dario Amodei, Tom Brown, Jack Clark, Jared Kaplan, Sam McCandlish, and Chris Olah. 2021.
\newblock A mathematical framework for transformer circuits.
\newblock \emph{Transformer Circuits Thread}.
\newblock Https://transformer-circuits.pub/2021/framework/index.html.

\bibitem[{Floridi and Chiriatti(2020)}]{Floridi2020-FLOGIN}
Luciano Floridi and Massimo Chiriatti. 2020.
\newblock \href {https://doi.org/10.1007/s11023-020-09548-1} {Gpt-3: Its nature, scope, limits, and consequences}.
\newblock \emph{Minds and Machines}, 30(4):681--694.

\bibitem[{Geva et~al.(2023)Geva, Bastings, Filippova, and Globerson}]{geva2023dissecting}
Mor Geva, Jasmijn Bastings, Katja Filippova, and Amir Globerson. 2023.
\newblock \href {http://arxiv.org/abs/2304.14767} {Dissecting recall of factual associations in auto-regressive language models}.

\bibitem[{Golovneva et~al.(2023)Golovneva, Chen, Poff, Corredor, Zettlemoyer, Fazel-Zarandi, and Celikyilmaz}]{golovneva2023roscoe}
Olga Golovneva, Moya Chen, Spencer Poff, Martin Corredor, Luke Zettlemoyer, Maryam Fazel-Zarandi, and Asli Celikyilmaz. 2023.
\newblock \href {http://arxiv.org/abs/2212.07919} {Roscoe: A suite of metrics for scoring step-by-step reasoning}.

\bibitem[{Goyal et~al.(2023)Goyal, Ji, Rawat, Menon, Kumar, and Nagarajan}]{goyal2023think}
Sachin Goyal, Ziwei Ji, Ankit~Singh Rawat, Aditya~Krishna Menon, Sanjiv Kumar, and Vaishnavh Nagarajan. 2023.
\newblock \href {http://arxiv.org/abs/2310.02226} {Think before you speak: Training language models with pause tokens}.

\bibitem[{Hagberg et~al.(2008)Hagberg, Schult, and Swart}]{hagberg2008exploring}
Aric~A. Hagberg, Daniel~A. Schult, and Pieter~J. Swart. 2008.
\newblock Exploring network structure, dynamics, and function using networkx.
\newblock In \emph{Proceedings of the 7th Python in Science Conference (SciPy2008)}, pages 11--15, Pasadena, CA USA.

\bibitem[{Han et~al.(2022)Han, Schoelkopf, Zhao, Qi, Riddell, Benson, Sun, Zubova, Qiao, Burtell, Peng, Fan, Liu, Wong, Sailor, Ni, Nan, Kasai, Yu, Zhang, Joty, Fabbri, Kryscinski, Lin, Xiong, and Radev}]{han2022folio}
Simeng Han, Hailey Schoelkopf, Yilun Zhao, Zhenting Qi, Martin Riddell, Luke Benson, Lucy Sun, Ekaterina Zubova, Yujie Qiao, Matthew Burtell, David Peng, Jonathan Fan, Yixin Liu, Brian Wong, Malcolm Sailor, Ansong Ni, Linyong Nan, Jungo Kasai, Tao Yu, Rui Zhang, Shafiq Joty, Alexander~R. Fabbri, Wojciech Kryscinski, Xi~Victoria Lin, Caiming Xiong, and Dragomir Radev. 2022.
\newblock \href {http://arxiv.org/abs/2209.00840} {Folio: Natural language reasoning with first-order logic}.

\bibitem[{Hou et~al.(2023)Hou, Li, Fei, Stolfo, Zhou, Zeng, Bosselut, and Sachan}]{hou2023mechanistic}
Yifan Hou, Jiaoda Li, Yu~Fei, Alessandro Stolfo, Wangchunshu Zhou, Guangtao Zeng, Antoine Bosselut, and Mrinmaya Sachan. 2023.
\newblock \href {http://arxiv.org/abs/2310.14491} {Towards a mechanistic interpretation of multi-step reasoning capabilities of language models}.

\bibitem[{Huang and Chang(2023)}]{reasoning_survey}
Jie Huang and Kevin Chen-Chuan Chang. 2023.
\newblock \href {https://doi.org/10.18653/v1/2023.findings-acl.67} {Towards reasoning in large language models: A survey}.
\newblock In \emph{Findings of the Association for Computational Linguistics: ACL 2023}, pages 1049--1065, Toronto, Canada. Association for Computational Linguistics.

\bibitem[{Huang et~al.(2022)Huang, Abbeel, Pathak, and Mordatch}]{hung2022zeroshotplanner}
Wenlong Huang, Pieter Abbeel, Deepak Pathak, and Igor Mordatch. 2022.
\newblock \href {https://proceedings.mlr.press/v162/huang22a.html} {Language models as zero-shot planners: Extracting actionable knowledge for embodied agents}.
\newblock In \emph{Proceedings of the 39th International Conference on Machine Learning}, volume 162 of \emph{Proceedings of Machine Learning Research}, pages 9118--9147. PMLR.

\bibitem[{Ivanitskiy et~al.(2023)Ivanitskiy, Spies, Räuker, Corlouer, Mathwin, Quirke, Rager, Shah, Valentine, Behn, Inoue, and Fung}]{ivanitskiy2023structured}
Michael~Igorevich Ivanitskiy, Alex~F. Spies, Tilman Räuker, Guillaume Corlouer, Chris Mathwin, Lucia Quirke, Can Rager, Rusheb Shah, Dan Valentine, Cecilia~Diniz Behn, Katsumi Inoue, and Samy~Wu Fung. 2023.
\newblock \href {http://arxiv.org/abs/2312.02566} {Structured world representations in maze-solving transformers}.

\bibitem[{Kojima et~al.(2023)Kojima, Gu, Reid, Matsuo, and Iwasawa}]{kojima2023large}
Takeshi Kojima, Shixiang~Shane Gu, Machel Reid, Yutaka Matsuo, and Yusuke Iwasawa. 2023.
\newblock \href {http://arxiv.org/abs/2205.11916} {Large language models are zero-shot reasoners}.

\bibitem[{Liu et~al.(2023)Liu, Ash, Goel, Krishnamurthy, and Zhang}]{liu2023transformers}
Bingbin Liu, Jordan~T. Ash, Surbhi Goel, Akshay Krishnamurthy, and Cyril Zhang. 2023.
\newblock \href {https://openreview.net/forum?id=De4FYqjFueZ} {Transformers learn shortcuts to automata}.
\newblock In \emph{The Eleventh International Conference on Learning Representations}.

\bibitem[{Marks and Tegmark(2023)}]{marks2023geometry}
Samuel Marks and Max Tegmark. 2023.
\newblock \href {http://arxiv.org/abs/2310.06824} {The geometry of truth: Emergent linear structure in large language model representations of true/false datasets}.

\bibitem[{Meng et~al.(2022)Meng, Bau, Andonian, and Belinkov}]{meng2022locating}
Kevin Meng, David Bau, Alex~J Andonian, and Yonatan Belinkov. 2022.
\newblock \href {https://openreview.net/forum?id=-h6WAS6eE4} {Locating and editing factual associations in {GPT}}.
\newblock In \emph{Advances in Neural Information Processing Systems}.

\bibitem[{Merrill et~al.(2021)Merrill, Goldberg, Schwartz, and Smith}]{merrill-etal-2021-provable}
William Merrill, Yoav Goldberg, Roy Schwartz, and Noah~A. Smith. 2021.
\newblock \href {https://doi.org/10.1162/tacl_a_00412} {Provable limitations of acquiring meaning from ungrounded form: What will future language models understand?}
\newblock \emph{Transactions of the Association for Computational Linguistics}, 9:1047--1060.

\bibitem[{Merrill et~al.(2022)Merrill, Sabharwal, and Smith}]{merrill-etal-2022-saturated}
William Merrill, Ashish Sabharwal, and Noah~A. Smith. 2022.
\newblock \href {https://doi.org/10.1162/tacl_a_00493} {Saturated transformers are constant-depth threshold circuits}.
\newblock \emph{Transactions of the Association for Computational Linguistics}, 10:843--856.

\bibitem[{Merullo et~al.(2024)Merullo, Eickhoff, and Pavlick}]{merullo2024circuit}
Jack Merullo, Carsten Eickhoff, and Ellie Pavlick. 2024.
\newblock \href {http://arxiv.org/abs/2310.08744} {Circuit component reuse across tasks in transformer language models}.

\bibitem[{Mikolov et~al.(2013)Mikolov, Yih, and Zweig}]{mikolov-etal-2013-linguistic}
Tomas Mikolov, Wen-tau Yih, and Geoffrey Zweig. 2013.
\newblock \href {https://aclanthology.org/N13-1090} {Linguistic regularities in continuous space word representations}.
\newblock In \emph{Proceedings of the 2013 Conference of the North {A}merican Chapter of the Association for Computational Linguistics: Human Language Technologies}, pages 746--751, Atlanta, Georgia. Association for Computational Linguistics.

\bibitem[{Mini et~al.(2023)Mini, Grietzer, Sharma, Meek, MacDiarmid, and Turner}]{mini2023understanding}
Ulisse Mini, Peli Grietzer, Mrinank Sharma, Austin Meek, Monte MacDiarmid, and Alexander~Matt Turner. 2023.
\newblock \href {http://arxiv.org/abs/2310.08043} {Understanding and controlling a maze-solving policy network}.

\bibitem[{Nanda and Bloom(2022)}]{nanda2022transformerlens}
Neel Nanda and Joseph Bloom. 2022.
\newblock Transformerlens.
\newblock \url{https://github.com/neelnanda-io/TransformerLens}.

\bibitem[{Nanda et~al.(2023{\natexlab{a}})Nanda, Chan, Lieberum, Smith, and Steinhardt}]{nanda2023progress}
Neel Nanda, Lawrence Chan, Tom Lieberum, Jess Smith, and Jacob Steinhardt. 2023{\natexlab{a}}.
\newblock \href {https://openreview.net/forum?id=9XFSbDPmdW} {Progress measures for grokking via mechanistic interpretability}.
\newblock In \emph{The Eleventh International Conference on Learning Representations}.

\bibitem[{Nanda et~al.(2023{\natexlab{b}})Nanda, Lee, and Wattenberg}]{nanda-etal-2023-emergent}
Neel Nanda, Andrew Lee, and Martin Wattenberg. 2023{\natexlab{b}}.
\newblock \href {https://doi.org/10.18653/v1/2023.blackboxnlp-1.2} {Emergent linear representations in world models of self-supervised sequence models}.
\newblock In \emph{Proceedings of the 6th BlackboxNLP Workshop: Analyzing and Interpreting Neural Networks for NLP}, pages 16--30, Singapore. Association for Computational Linguistics.

\bibitem[{Nanda et~al.(2023{\natexlab{c}})Nanda, Rajamanoharan, Kramar, and Shah}]{nanda2023factfinding}
Neel Nanda, Senthooran Rajamanoharan, Janos Kramar, and Rohin Shah. 2023{\natexlab{c}}.
\newblock \href {https://www.alignmentforum.org/posts/iGuwZTHWb6DFY3sKB/fact-finding-attempting-to-reverse-engineer-factual-recall} {Fact finding: Attempting to reverse-engineer factual recall on the neuron level}.

\bibitem[{Olsson et~al.(2022)Olsson, Elhage, Nanda, Joseph, DasSarma, Henighan, Mann, Askell, Bai, Chen, Conerly, Drain, Ganguli, Hatfield-Dodds, Hernandez, Johnston, Jones, Kernion, Lovitt, Ndousse, Amodei, Brown, Clark, Kaplan, McCandlish, and Olah}]{olsson2022incontext}
Catherine Olsson, Nelson Elhage, Neel Nanda, Nicholas Joseph, Nova DasSarma, Tom Henighan, Ben Mann, Amanda Askell, Yuntao Bai, Anna Chen, Tom Conerly, Dawn Drain, Deep Ganguli, Zac Hatfield-Dodds, Danny Hernandez, Scott Johnston, Andy Jones, Jackson Kernion, Liane Lovitt, Kamal Ndousse, Dario Amodei, Tom Brown, Jack Clark, Jared Kaplan, Sam McCandlish, and Chris Olah. 2022.
\newblock \href {http://arxiv.org/abs/2209.11895} {In-context learning and induction heads}.

\bibitem[{Pérez et~al.(2021)Pérez, Barceló, and Marinkovic}]{perez2021attention}
Jorge Pérez, Pablo Barceló, and Javier Marinkovic. 2021.
\newblock \href {http://jmlr.org/papers/v22/20-302.html} {Attention is turing-complete}.
\newblock \emph{Journal of Machine Learning Research}, 22(75):1--35.

\bibitem[{Russell and Norvig(2009)}]{Russell2009-ht}
Stuart Russell and Peter Norvig. 2009.
\newblock \emph{Artificial intelligence}, 3 edition.
\newblock Pearson, Upper Saddle River, New Jersey.

\bibitem[{Räuker et~al.(2023)Räuker, Ho, Casper, and Hadfield-Menell}]{räuker2023transparent}
Tilman Räuker, Anson Ho, Stephen Casper, and Dylan Hadfield-Menell. 2023.
\newblock \href {http://arxiv.org/abs/2207.13243} {Toward transparent ai: A survey on interpreting the inner structures of deep neural networks}.

\bibitem[{Saparov and He(2023)}]{saparov2023language}
Abulhair Saparov and He~He. 2023.
\newblock \href {https://openreview.net/forum?id=qFVVBzXxR2V} {Language models are greedy reasoners: A systematic formal analysis of chain-of-thought}.
\newblock In \emph{The Eleventh International Conference on Learning Representations}.

\bibitem[{Stolfo et~al.(2023)Stolfo, Belinkov, and Sachan}]{stolfo-etal-2023-mechanistic}
Alessandro Stolfo, Yonatan Belinkov, and Mrinmaya Sachan. 2023.
\newblock \href {https://doi.org/10.18653/v1/2023.emnlp-main.435} {A mechanistic interpretation of arithmetic reasoning in language models using causal mediation analysis}.
\newblock In \emph{Proceedings of the 2023 Conference on Empirical Methods in Natural Language Processing}, pages 7035--7052, Singapore. Association for Computational Linguistics.

\bibitem[{Strobl et~al.(2023)Strobl, Merrill, Weiss, Chiang, and Angluin}]{strobl2023transformers}
Lena Strobl, William Merrill, Gail Weiss, David Chiang, and Dana Angluin. 2023.
\newblock \href {http://arxiv.org/abs/2311.00208} {Transformers as recognizers of formal languages: A survey on expressivity}.

\bibitem[{Syed et~al.(2023)Syed, Rager, and Conmy}]{syed2023attribution}
Aaquib Syed, Can Rager, and Arthur Conmy. 2023.
\newblock \href {http://arxiv.org/abs/2310.10348} {Attribution patching outperforms automated circuit discovery}.

\bibitem[{Tafjord et~al.(2021)Tafjord, Dalvi, and Clark}]{tafjord-etal-2021-proofwriter}
Oyvind Tafjord, Bhavana Dalvi, and Peter Clark. 2021.
\newblock \href {https://doi.org/10.18653/v1/2021.findings-acl.317} {{P}roof{W}riter: Generating implications, proofs, and abductive statements over natural language}.
\newblock In \emph{Findings of the Association for Computational Linguistics: ACL-IJCNLP 2021}, pages 3621--3634, Online. Association for Computational Linguistics.

\bibitem[{Tigges et~al.(2023)Tigges, Hollinsworth, Geiger, and Nanda}]{tigges2023linear}
Curt Tigges, Oskar~John Hollinsworth, Atticus Geiger, and Neel Nanda. 2023.
\newblock \href {http://arxiv.org/abs/2310.15154} {Linear representations of sentiment in large language models}.

\bibitem[{Valmeekam et~al.(2023)Valmeekam, Marquez, Olmo, Sreedharan, and Kambhampati}]{valmeekam2023planbench}
Karthik Valmeekam, Matthew Marquez, Alberto Olmo, Sarath Sreedharan, and Subbarao Kambhampati. 2023.
\newblock \href {https://openreview.net/forum?id=YXogl4uQUO} {Planbench: An extensible benchmark for evaluating large language models on planning and reasoning about change}.
\newblock In \emph{Thirty-seventh Conference on Neural Information Processing Systems Datasets and Benchmarks Track}.

\bibitem[{Variengien and Winsor(2023)}]{variengien2023look}
Alexandre Variengien and Eric Winsor. 2023.
\newblock \href {http://arxiv.org/abs/2312.10091} {Look before you leap: A universal emergent decomposition of retrieval tasks in language models}.

\bibitem[{Vaswani et~al.(2017)Vaswani, Shazeer, Parmar, Uszkoreit, Jones, Gomez, Kaiser, and Polosukhin}]{vaswani2017attention}
Ashish Vaswani, Noam Shazeer, Niki Parmar, Jakob Uszkoreit, Llion Jones, Aidan~N. Gomez, Lukasz Kaiser, and Illia Polosukhin. 2017.
\newblock \href {http://arxiv.org/abs/1706.03762} {Attention is all you need}.

\bibitem[{Vig and Belinkov(2019)}]{vig-belinkov-2019-analyzing}
Jesse Vig and Yonatan Belinkov. 2019.
\newblock \href {https://doi.org/10.18653/v1/W19-4808} {Analyzing the structure of attention in a transformer language model}.
\newblock In \emph{Proceedings of the 2019 ACL Workshop BlackboxNLP: Analyzing and Interpreting Neural Networks for NLP}, pages 63--76, Florence, Italy. Association for Computational Linguistics.

\bibitem[{Vig et~al.(2020)Vig, Gehrmann, Belinkov, Qian, Nevo, Singer, and Shieber}]{NEURIPS2020_92650b2e}
Jesse Vig, Sebastian Gehrmann, Yonatan Belinkov, Sharon Qian, Daniel Nevo, Yaron Singer, and Stuart Shieber. 2020.
\newblock \href {https://proceedings.neurips.cc/paper_files/paper/2020/file/92650b2e92217715fe312e6fa7b90d82-Paper.pdf} {Investigating gender bias in language models using causal mediation analysis}.
\newblock In \emph{Advances in Neural Information Processing Systems}, volume~33, pages 12388--12401. Curran Associates, Inc.

\bibitem[{Wang et~al.(2023{\natexlab{a}})Wang, Min, Deng, Shen, Wu, Zettlemoyer, and Sun}]{wang-etal-2023-towards}
Boshi Wang, Sewon Min, Xiang Deng, Jiaming Shen, You Wu, Luke Zettlemoyer, and Huan Sun. 2023{\natexlab{a}}.
\newblock \href {https://doi.org/10.18653/v1/2023.acl-long.153} {Towards understanding chain-of-thought prompting: An empirical study of what matters}.
\newblock In \emph{Proceedings of the 61st Annual Meeting of the Association for Computational Linguistics (Volume 1: Long Papers)}, pages 2717--2739, Toronto, Canada. Association for Computational Linguistics.

\bibitem[{Wang et~al.(2023{\natexlab{b}})Wang, Variengien, Conmy, Shlegeris, and Steinhardt}]{wang2023interpretability}
Kevin~Ro Wang, Alexandre Variengien, Arthur Conmy, Buck Shlegeris, and Jacob Steinhardt. 2023{\natexlab{b}}.
\newblock \href {https://openreview.net/forum?id=NpsVSN6o4ul} {Interpretability in the wild: a circuit for indirect object identification in {GPT}-2 small}.
\newblock In \emph{The Eleventh International Conference on Learning Representations}.

\bibitem[{Wei et~al.(2022)Wei, Wang, Schuurmans, Bosma, Ichter, Xia, Chi, Le, and Zhou}]{wei2022chain}
Jason Wei, Xuezhi Wang, Dale Schuurmans, Maarten Bosma, Brian Ichter, Fei Xia, Ed~H. Chi, Quoc~V Le, and Denny Zhou. 2022.
\newblock \href {https://openreview.net/forum?id=_VjQlMeSB_J} {Chain of thought prompting elicits reasoning in large language models}.
\newblock In \emph{Advances in Neural Information Processing Systems}.

\bibitem[{Yun et~al.(2020)Yun, Bhojanapalli, Rawat, Reddi, and Kumar}]{Yun2020Are}
Chulhee Yun, Srinadh Bhojanapalli, Ankit~Singh Rawat, Sashank Reddi, and Sanjiv Kumar. 2020.
\newblock \href {https://openreview.net/forum?id=ByxRM0Ntvr} {Are transformers universal approximators of sequence-to-sequence functions?}
\newblock In \emph{International Conference on Learning Representations}.

\bibitem[{Zhang et~al.(2024)Zhang, Tigges, Zhang, Biderman, Raginsky, and Ringer}]{zhang2024transformerbased}
Dylan Zhang, Curt Tigges, Zory Zhang, Stella Biderman, Maxim Raginsky, and Talia Ringer. 2024.
\newblock \href {http://arxiv.org/abs/2401.12947} {Transformer-based models are not yet perfect at learning to emulate structural recursion}.

\bibitem[{Zhong et~al.(2023)Zhong, Liu, Tegmark, and Andreas}]{zhong2023clock}
Ziqian Zhong, Ziming Liu, Max Tegmark, and Jacob Andreas. 2023.
\newblock \href {http://arxiv.org/abs/2306.17844} {The clock and the pizza: Two stories in mechanistic explanation of neural networks}.

\end{thebibliography}

\appendix

\label{sec:appendix}

\section{Extended Discussion of Mechanistic Interpretability}
\label{app:mechanistic-interpretability}
Our work contributes to the emerging field of mechanistic interpretability, which seeks to reverse-engineer the internal mechanisms of neural networks into human-understandable algorithms~\citep{elhage2021mathematical}. 
To this end, the model is considered a causal graph~\citep{meng2022locating}, with the intent of finding interpretable and meaningful subgraphs (circuits) that are responsible for solving the task in question, such as understanding a circuit implementing modular addition~\citep{nanda2023progress}, or understanding a circuit responsible for indirect object identification~\citep{wang2023interpretability}.
Similar works studied how models perform in-context learning using induction heads, as found by~\citet{olsson2022incontext}, or how models perform retrieval tasks using an internal modular decomposition as shown in~\citet{variengien2023look}. 
Moreover,~\citet{merullo2024circuit} provide evidence that certain mechanisms are reused across different tasks, indicating that models can use similar mechanisms to solve different tasks. 

However, these studies of specific mechanisms and circuits in language models required a lot of manual effort.
In addition, \citet{zhong2023clock} demonstrated that small changes in hyperparameters and initializations can lead to the emergence of qualitatively different algorithms.
To address this, some researchers attempted to automate the process of finding these circuits using methods such as automated circuit discovery~\citep{conmy2023automated} or edge attribution patching~\citep{syed2023attribution}.

Similar research investigates how neural networks process and store information.
For example,~\citet{geva2023dissecting} explored how models recall factual information, and \citet{meng2022locating} studied how this information can be localized and edited without re-training. 
This also includes works on understanding internal representations. 
In this context, there is an ongoing debate about the linear representation hypothesis~\cite{mikolov-etal-2013-linguistic}, which is the idea that high-level concepts are represented \textit{linearly} as directions in the representation space.
Here, a model is considered to linearly represent a concept, if it can be probed from the activations with a linear model. 
For example, \citet{nanda-etal-2023-emergent} and \citet{mini2023understanding} studied the internal representations of language models trained on a board game and found that it develops a representation of the current board state that can be extracted using linear probes.
Importantly, they find that whether the representation can be extracted using a linear model depends on the encoding of the target label.
Other linear representations have been found in language models; e.g.\ \citet{tigges2023linear} find a linear representation of sentiment in text and~\citet{marks2023geometry} find a linear representation of the truth value of input text.
\citet{ivanitskiy2023structured} explored a transformer trained to perform a maze-solving task and found that they can successfully extract boundaries of the maze using linear probes.
They also observed attention heads that attend to positions in the maze that are one step away from the current position. %

\section{Experimental Setup}
\label{app:experimental-setup}

\subsection{Implementation and Computing}
All experiments were carried out on a single NVIDIA RTX A6000 GPU.
The total computation time for training the transformer model was less than 24 hours. 
To generate the trees, we used \texttt{networkx}~\citep{hagberg2008exploring}.
For training and execution of all experiments, we used \texttt{TransformerLens}~\citep{nanda2022transformerlens}.
For details on the model and training configuration, see Tables~\ref{tab:model-configuration} and~\ref{tab:hyperparameters}.

\begin{table}[!h]
    \centering
    \caption{Model Configuration}
    \begin{tabularx}{\columnwidth}{X Y}
    \toprule
    \textbf{Parameter} & \textbf{Value} \\
    \midrule
    \# of layers & 6 \\
    \# of heads & 1 \\
    Residual Stream dim. & 128 \\
    Attention Head dim. & 128 \\
    Feed-Forward dim. & 512 \\
    Activation Function & \texttt{gelu} \\
    Vocabulary Size & 35 \\
    Context Size & 63 \\
    \bottomrule
    \end{tabularx}
    \label{tab:model-configuration}
\end{table}

\begin{table}[!h]
    \centering
    \caption{Training Configuration}
    \begin{tabularx}{\columnwidth}{X Y}
    \toprule
    \textbf{Parameter} & \textbf{Value} \\
    \midrule
    Learning Rate & 1e-3 \\
    Optimizer & \texttt{AdamW} \\
    Batch Size & 64 \\
    Betas & (0.9, 0.99) \\
    Weight Decay & 0.01 \\
    \bottomrule
    \end{tabularx}
    \label{tab:hyperparameters}
\end{table}

\subsection{Size of Training Set}
\label{app:training-samples}
Our dataset consists of 150,000 randomly generated examples, each including a labeled binary tree with 16 nodes.
The number of possible \textit{unlabeled} binary trees with $n+1$ nodes is given by the $n$-th Catalan number~\citep{Catalan1838}:
$$C(n) = \frac{(2n)!}{(n + 1)! \cdot n!}$$
When considering \textit{labeled} binary trees, this number grows to $(n+1)! \cdot C(n)$ unique trees.
This suggests that memorization is infeasible, and generalization is required for meaningful performance.

\section{Attention Head Composition}
\label{app:head-composition}
In this section, we perform additional experiments to verify that \texttt{L1.H1} performs edge token concatenation, and \texttt{L2.H1} is a deduction head.
\citet{elhage2021mathematical} show that transformers can learn induction heads in two different ways involving different compositions: the K-composition, where the $W_K$ of the head reads from the output of the previous head, and V-composition, where the $W_V$ of the head reads from the output of the previous head.
Our findings suggest that the deduction heads we found in our model are a result of K-composition. 
In the following subsections, we adopt the conventions of~\cite{elhage2021mathematical}.

\begin{figure}[t]
    \centering
    \vspace{-0.6cm}
    \hspace*{-0.7cm}\includegraphics[width=1.2\columnwidth]{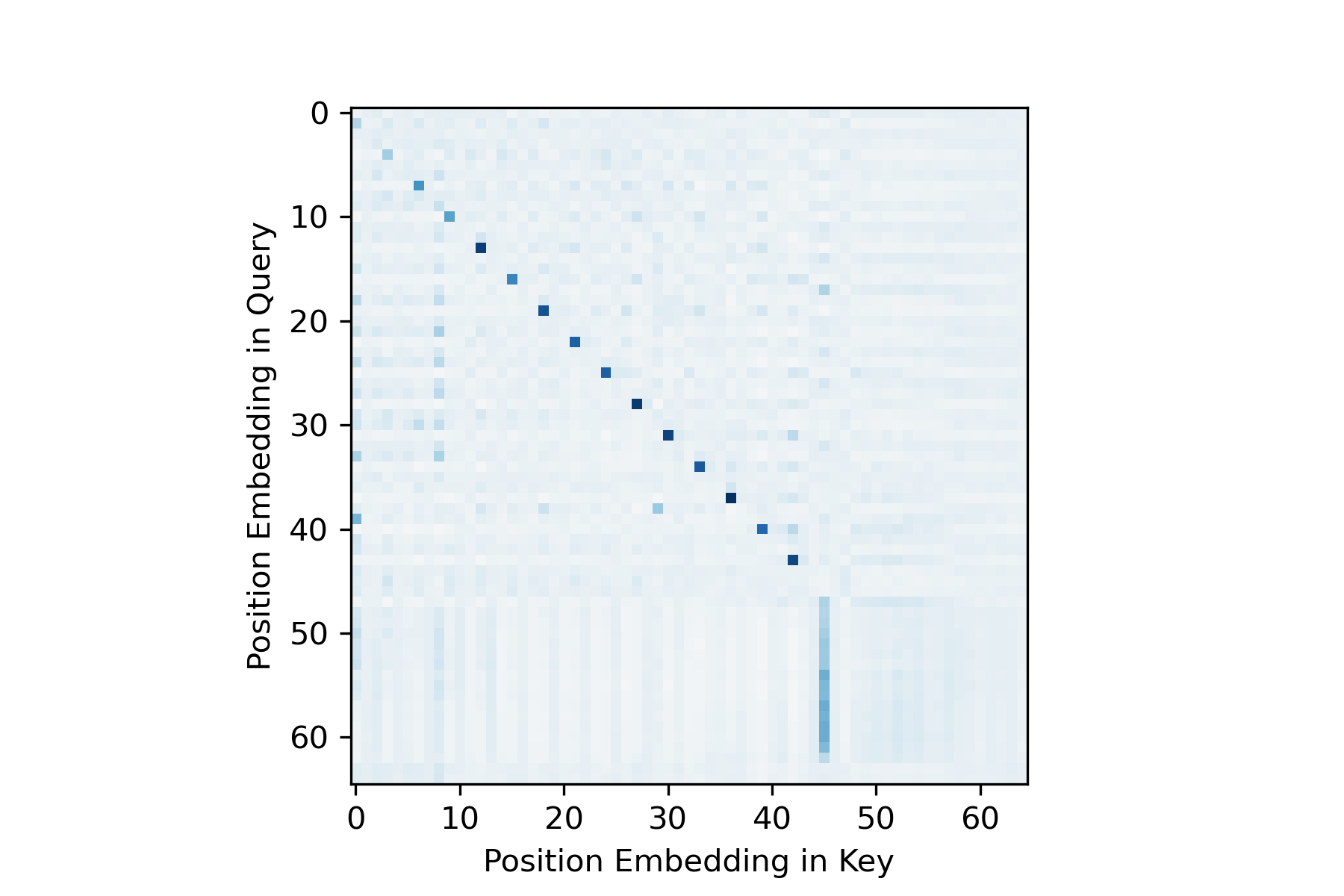}
    \caption{Visualization of $M_{QK}^0$}
    \label{fig:qk}
    \vspace{-0.2cm}
\end{figure}

\paragraph{Layer 1 - Edge Token Concatenation Head} 
\hspace{1pt}
\texttt{L1.H1} serves as a variation of a previous token head studied in~\citet{elhage2021mathematical}, effectively transferring source node information to the corresponding target node in each edge.
This is captured in the QK-circuit: 
$$M^0_{QK} = W_P\, W_{QK}^0\, W_P^T$$
$M^0_{QK}$ (see Figure~\ref{fig:qk}) shows that the attention values are maximized when the query vector corresponds to the position embedding of an incoming node, and the key vector corresponds to the position embedding of the immediately preceding outgoing node. 
Then, following the goal node at position 45, the head persistently attends to the goal position. 

\begin{figure}[t]
    \centering
    \vspace{-0.6cm}
    \hspace*{-0.7cm}\includegraphics[width=1.2\columnwidth]{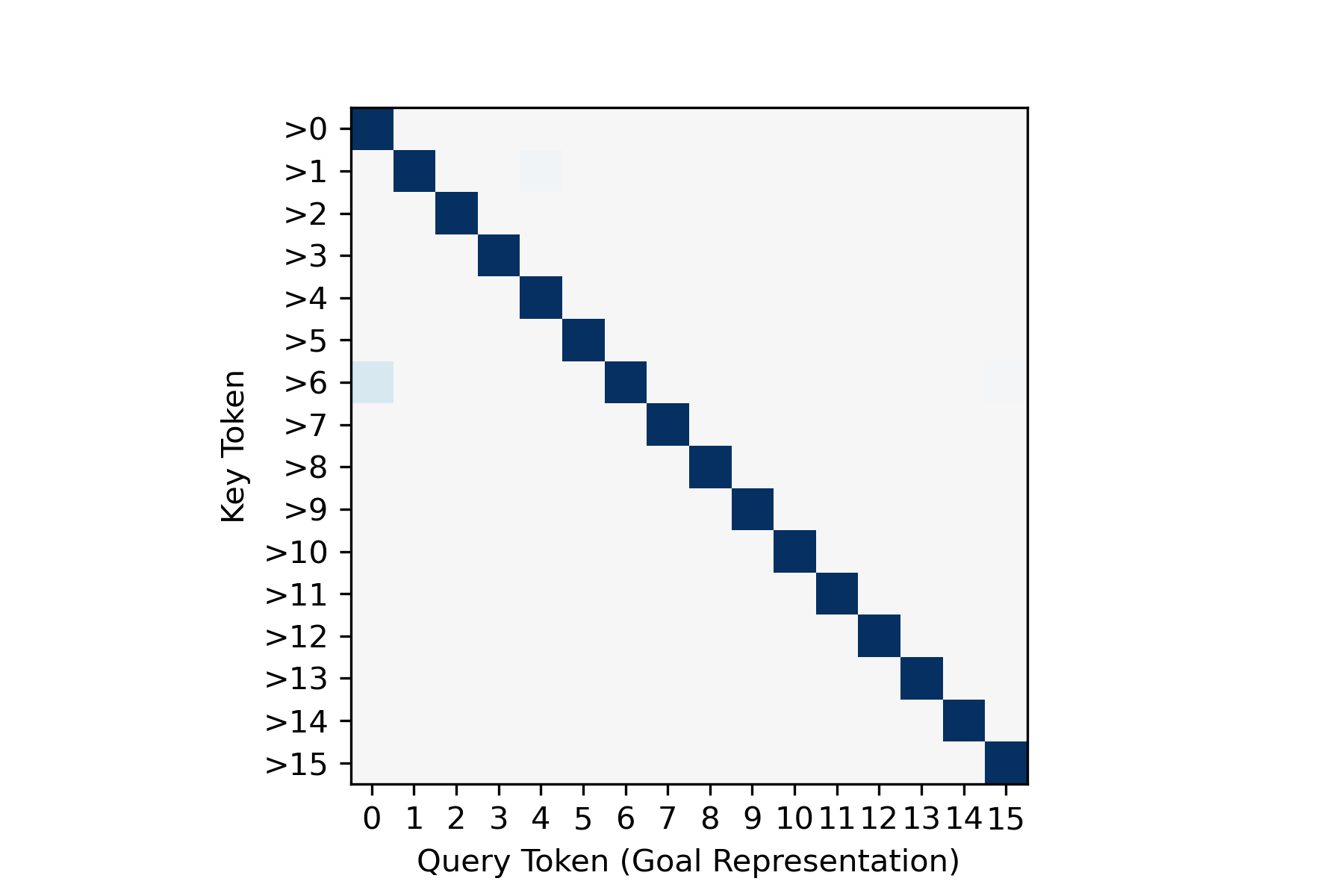}
    \caption{Visualization of a subset of $M_{QK}^1$ showing interactions between source and target node tokens.}
    \label{fig:k-comp}
    \vspace{-0.2cm}
\end{figure}

\paragraph{Layer 2 - Deduction Head}
\texttt{L2.H1} is a deduction head, which attends to the target node that matches the goal at positions in the path.
It then moves information about the source node of that edge into the last position in the window. 
It can be viewed as a reverse induction head~\citep{olsson2022incontext} that uses a K-composition~\citep{elhage2021mathematical} to map a sequence \token{A} \token{B} \ldots \token{A} $\rightarrow$ \token{B}, where \token{B} represents target nodes and \token{A} represents source nodes.
This can again be verified by looking at the QK-circuit:  
$$M^1_{QK} = (\text{MLP}^0(W^0_{OV} W_E) +  W^0_{OV} W_E) W^1_{QK} W_E^T$$
This matrix shows the interactions of the embedding of the source and target tokens at layer 2 (see Figure~\ref{fig:k-comp}).
Our analysis is complicated by the fact that our model is not attention-only, as attention heads can compose with each other through the MLP, which makes similar analyses in later layers of the model intractable.
However, our causal scrubbing results provide evidence that the attention heads in the subsequent layers implement a similar mechanism to \texttt{L2.H1}, but use the output of the previous layer's attention head to backward chain further up the tree.

\section{Register Tokens and Subgoals}
\label{app:register-tokens}
In this section, we provide more details on the role of the register tokens in our model.
From each register token position, the model attends to a random node in the context and starts backward-chaining from that node.
The initial selection of a node by the register token can be viewed as identifying a \textit{subgoal}, from which the model can perform backward chaining. 
This precomputation occurs before the actual goal is specified and occurs fully parallel to the main backward-chaining mechanism.
These findings hint that transformers may exhibit an inductive bias towards learning highly parallelized algorithms when trained to perform search, planning, or reasoning.

\begin{figure}[!h]
    \centering
    \includegraphics[width=\linewidth]{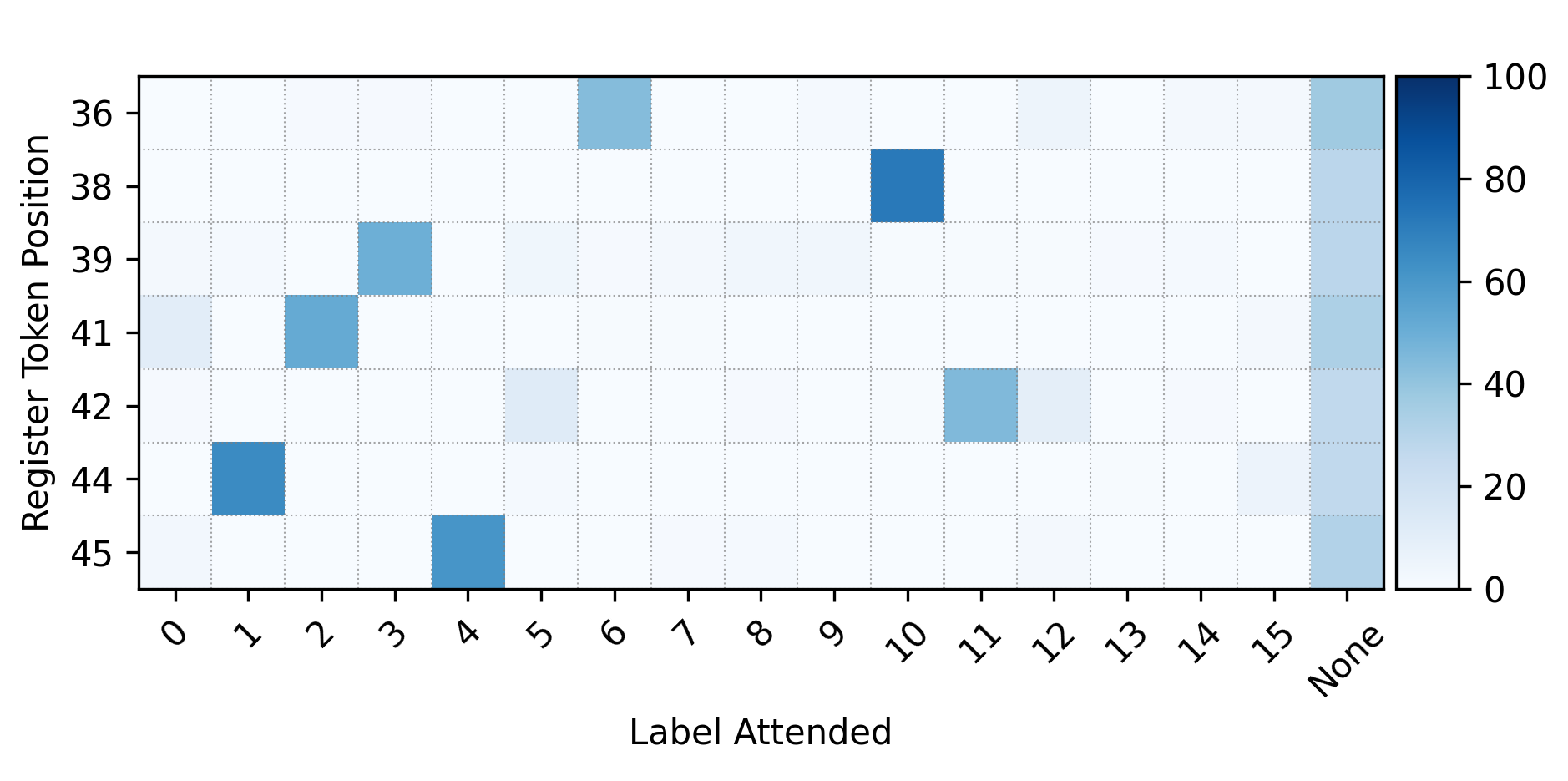} %
    \vspace{-0.4cm}
    \caption{Preferences in Subgoal Selection: Ratio of register tokens attending to different subgoals aggregated across 1000 trees. We consider a register token to select a subgoal based on an attention threshold of 0.3.}
    \label{fig:register-tokens-empirical}
 \end{figure}

To see whether there is any structure in the selection of subgoals, we empirically study which node the register tokens select as subgoals across 1000 samples. 
The results are illustrated in Figures~\ref{fig:register-tokens-empirical}. 

\begin{figure}[!h]
 \centering
 \includegraphics[width=\linewidth]{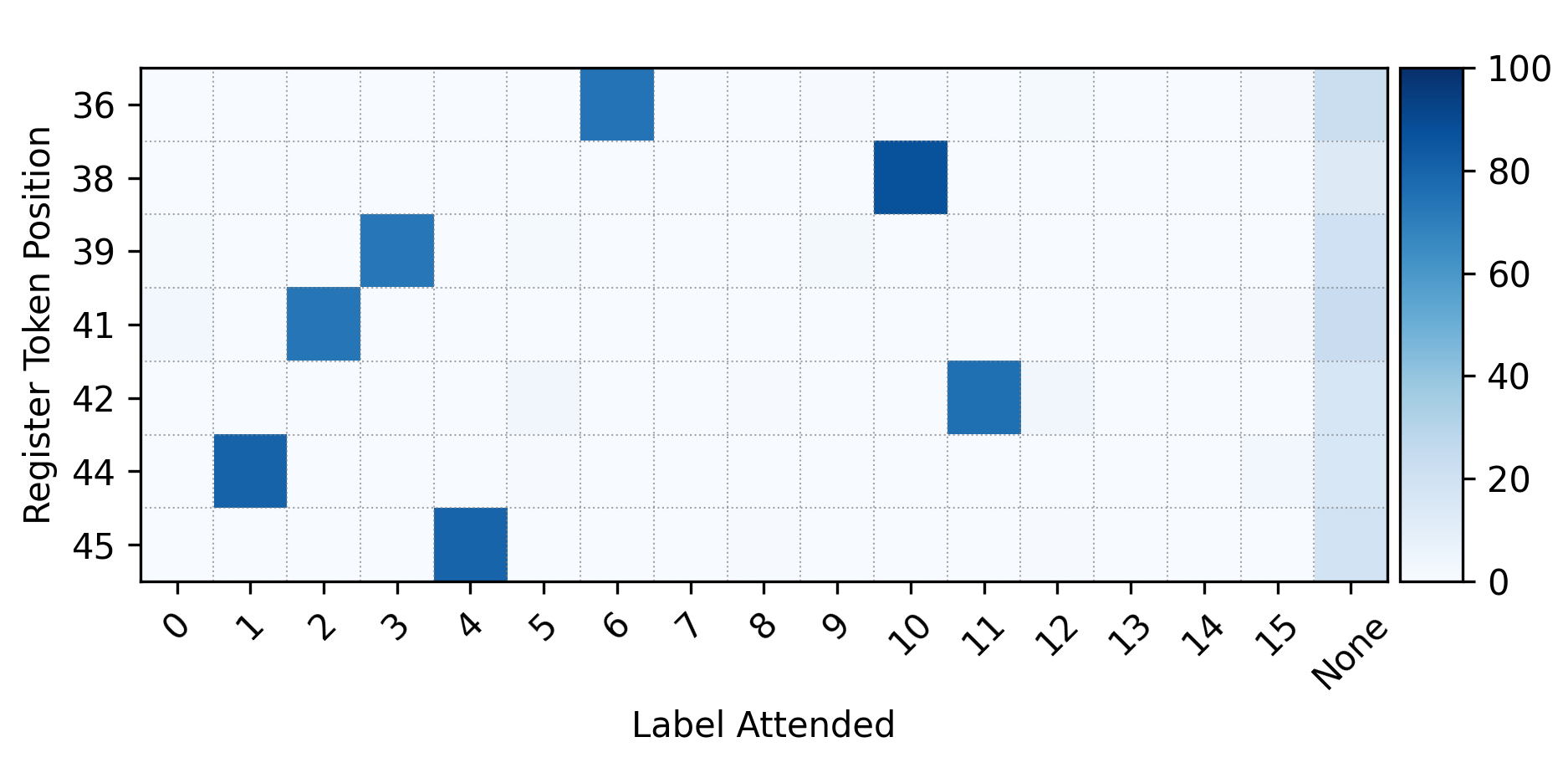} %
 \vspace{-0.4cm}
 \caption{Preferences in Subgoal Selection: Ratio of register tokens attending to different subgoals aggregated across 1000 trees where the node it attends to most often is not a leaf node of the tree and occurs before the register token. We consider a register token to select a subgoal based on an attention threshold of 0.3.}
 \label{fig:subgoal-selection-empirical-noleaf}
\end{figure}

We observe that the model usually attends to the same tokens, e.g.\ position 36 attends to token \token{6} most of the time.
However, we observe an interesting dynamic in which the register token selects a different subgoal in two cases: 
\begin{enumerate}
    \item If the node doesn't occur before the register token position, it cannot attend to it due to causal masking.
    \item If the node is a leaf node of the tree since it doesn't have a corresponding source token to attend to. 
\end{enumerate}
To validate this, we again examine the probability of the model selecting subgoals in trees where the most common subgoal occurs before the register token position and is not a leaf node (see Figure~\ref{fig:subgoal-selection-empirical-noleaf}).

Further exploration reveals that the subgoals selected by each register token position can be somewhat understood through an examination of the embedding matrices. 
We evaluate a selection of seven register token positions that are used on several different examples and show their preferred subgoals.
By composing the embedding and position embedding matrices with the QK-circuit of the first layer's attention head, we define $R_P$ as: 
$$R_P = W_P W^1_{QK} W_E^T$$
where $W_E$ is the embedding matrix, $W_P$ is the position embedding matrix, and $W^1_E, W^1_Q$ are the key and query projection matrices of \texttt{L1.H1}.
This explains how the model selects subgoals, by having the key for each positional embedding of a register token match with some specific source node token.

 \begin{figure}[!h]
    \centering
    \vspace{-0.3cm}
    \includegraphics[width=\linewidth]{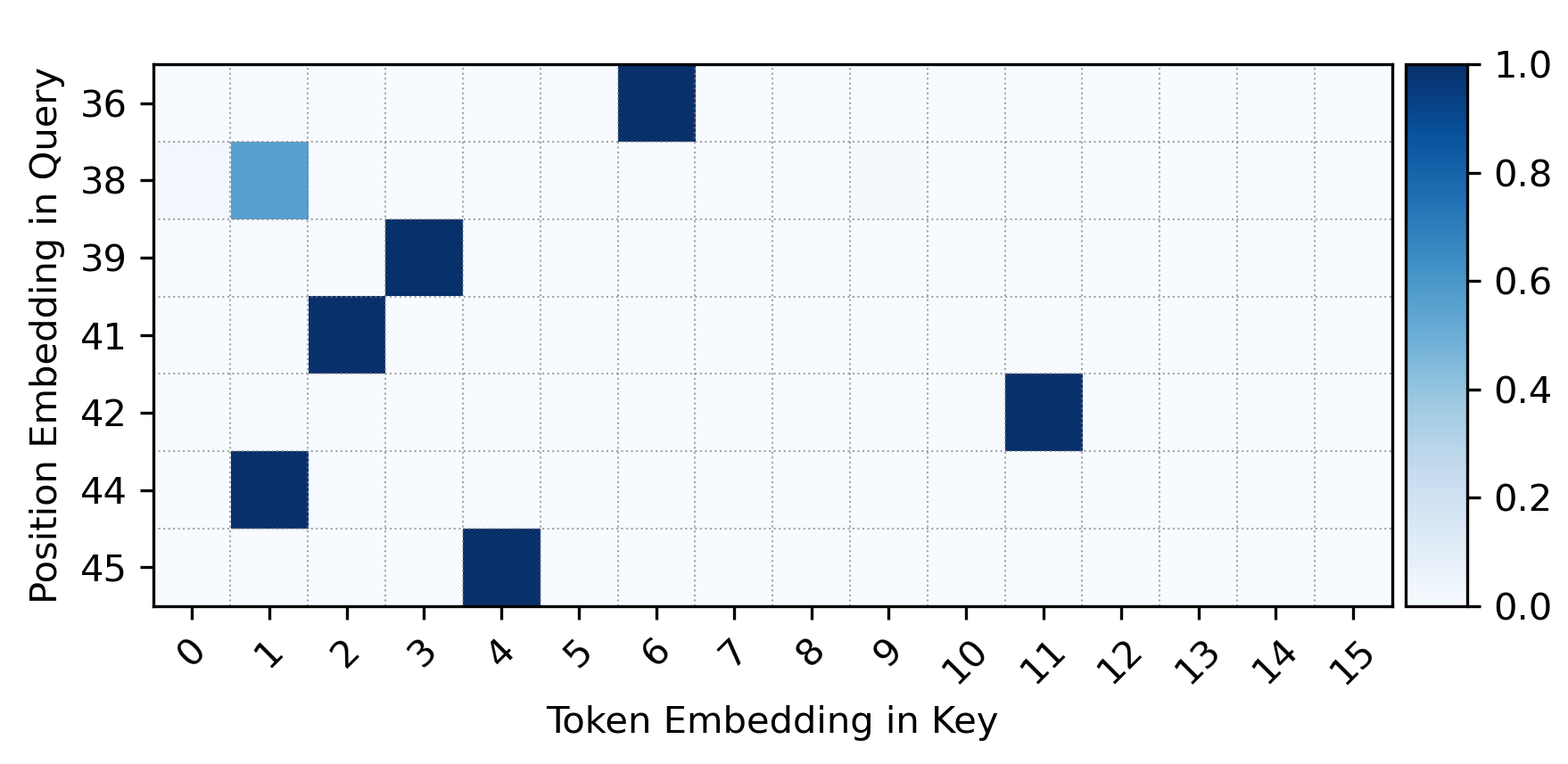} %
    \caption{Plot of $R_P$}
    \label{fig:register-tokens-im1}
 \end{figure}

\section{Attention Patterns}
\label{app:attention-patterns}
\begin{figure*}[!t]
    \centering
    \includegraphics{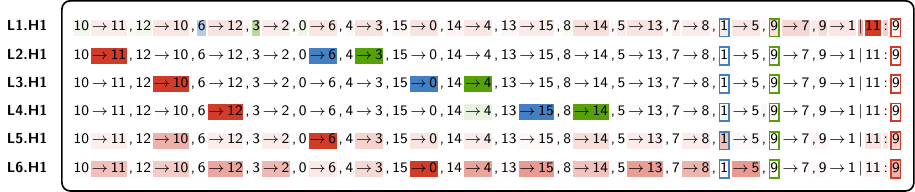}
    \caption{Visualization of multi-layer attention patterns on an example input, similar to Figure~\ref{fig:attention-patterns}. We show the attention from three different positions: \textcolor{red!199}{path position}, \textcolor{blue!199}{register token at position 39}, and \textcolor{green!199}{register token at position 45}.}
    \label{fig:attention-patterns-1}
\end{figure*}

\begin{figure*}[!t]
    \centering
    \includegraphics{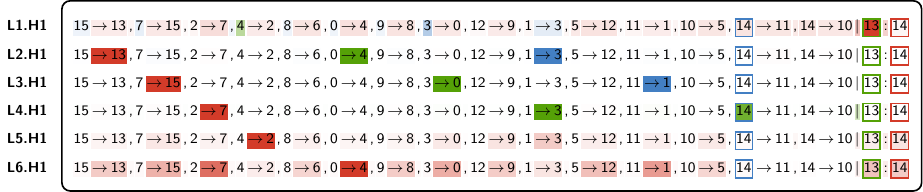}
    \caption{Visualization of multi-layer attention patterns on an example input, similar to Figure~\ref{fig:attention-patterns}. We show the attention from three different positions: the \textcolor{red!199}{path position} and \textcolor{blue!199}{register token at position 41}}
    \label{fig:attention-patterns-2}
\end{figure*}

\begin{figure*}[!t]
    \centering
    \includegraphics{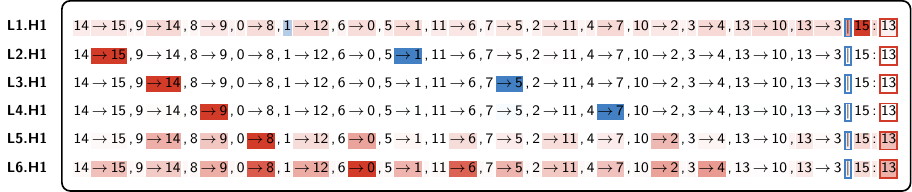}
    \caption{Visualization of multi-layer attention patterns on an example input, similar to Figure~\ref{fig:attention-patterns}. We show the attention from three different positions: the \textcolor{red!199}{path position} and \textcolor{blue!199}{register token at position 36}}
    \label{fig:attention-patterns-3}
\end{figure*}

\begin{figure*}[!t]
    \centering
    \includegraphics{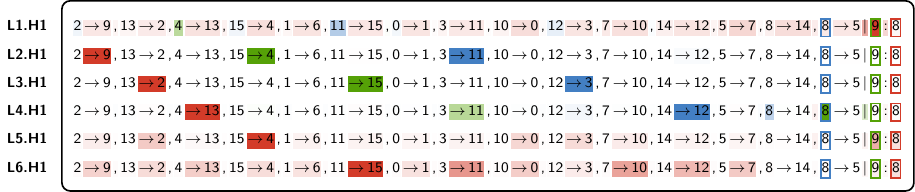}
    \caption{Visualization of multi-layer attention patterns on an example input, similar to Figure~\ref{fig:attention-patterns}. We show the attention from three different positions: \textcolor{red!199}{path position}, \textcolor{blue!199}{register token at position 41}, and \textcolor{green!199}{register token at position 42}.}
    \label{fig:attention-patterns-4}
\end{figure*}

Here, we visualize attention patterns of our model on a few example inputs (see Figures~\ref{fig:attention-patterns-1} to \ref{fig:attention-patterns-4}), in addition to the example illustrated in Figure~\ref{fig:attention-patterns}, to provide intuition for the backward chaining mechanism. 
In each example, we highlight the attention from the final token position and the register tokens that are causally relevant for the prediction.
To determine these, we use attention knockout~\cite{geva2023dissecting} on the register token positions.
This prevents the final token from attending to register tokens by zero-ing out the attention weights.
This allows us to test whether critical information propagates from them. 
More formally, let $a, b \in [1, N]$ be two positions such that $a <= b$, we block $\mathbf{x}_b^l$ from attending to $\mathbf{x}_a^\ell$ at layer $\ell < L$ by updating the attention weights to that layer:
$$M_{ab}^{l+1} = -\infty \forall j\in[1,H]$$
Thus, this restricts the source position from obtaining information from the target position, at that particular layer.
To avoid information leakage across positions, we block attention edges in multiple layers rather than a single one.
Specifically, we block attention to the register tokens in the final two layers of the model.

\clearpage

\clearpage
\section{Additional Experiment on the One-Step Lookahead}
\label{app:mechanism-children}

\begin{figure}[!h]
    \centering
    \includegraphics{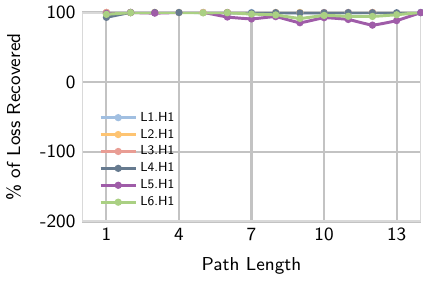}
    \caption{To test the impact of the one-step lookahead, we replicate the causal srubbing experiment from Section~\ref{cha:backward-chaining} but add additional constraints to our resampling scheme. 
    We find that we can recover most of the model performance across the full training distribution. 
    }
    \label{fig:backward-chaining-causal-scrubbing-with-fallback-mechanism}
\end{figure}

To evaluate the impact of the one-step lookahead, we perform causal scrubbing that incorporates the described mechanism.
We reuse the experimental setup from Section~\ref{cha:backward-chaining} but add additional constraints to our resampling scheme.
Specifically, we avoid resampling the contributions of the target node and register token positions through the attention heads of the last two layers.
We visualize the results in Figure~\ref{fig:backward-chaining-causal-scrubbing-with-fallback-mechanism}.

\section{Tuned Lens}
\label{app:tuned-lens}
In this section, we provide an additional piece of evidence in favor of the existence of the backward chaining mechanism.
To understand how the predictions of a transformer are built layer-by-layer,~\citet{belrose2023eliciting} develop the Tuned Lens, a method that involves training a linear model to translate the activations from an intermediate layer directly to the input of the unembedding layer.
 
\begin{figure*}[!h]
     \centering
     \includegraphics[width=0.9\textwidth]{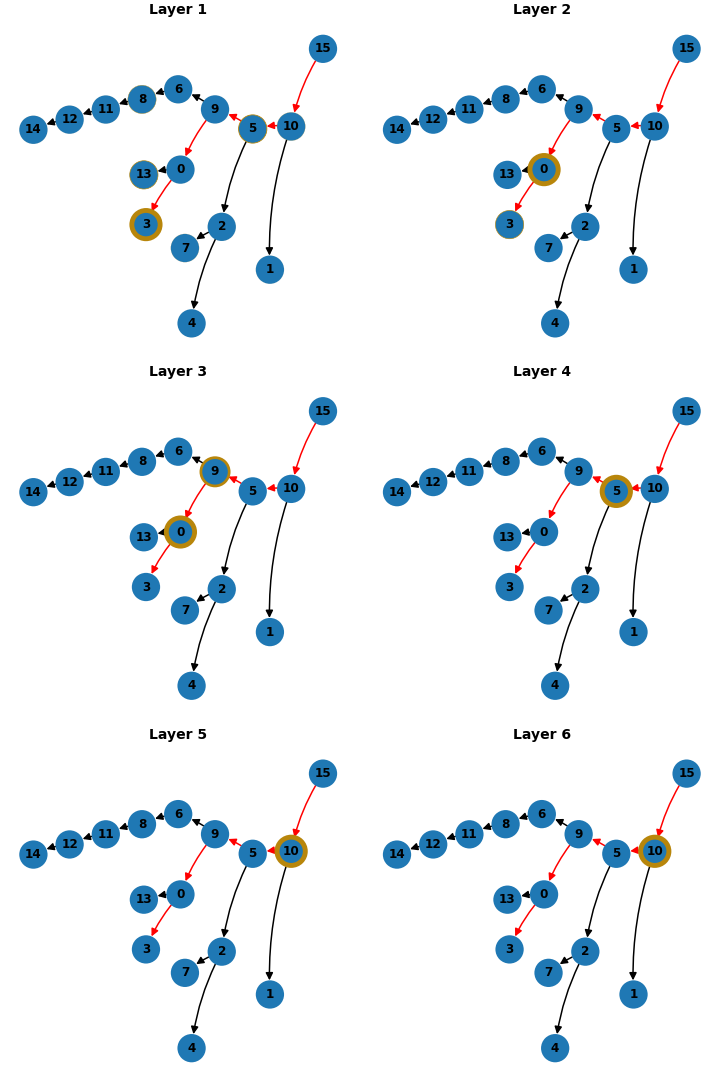} %
     \caption{Example 1 (Path Length 5): Results of a linear transformation to predict the next step based on the residual stream activations after each layer, projected onto the tree structure. The yellow border highlights the current best prediction(s) of the linear transformation.}
     \label{fig:tl1}
 \end{figure*}

\begin{figure*}[!h]
     \centering
     \includegraphics[width=0.9\textwidth]{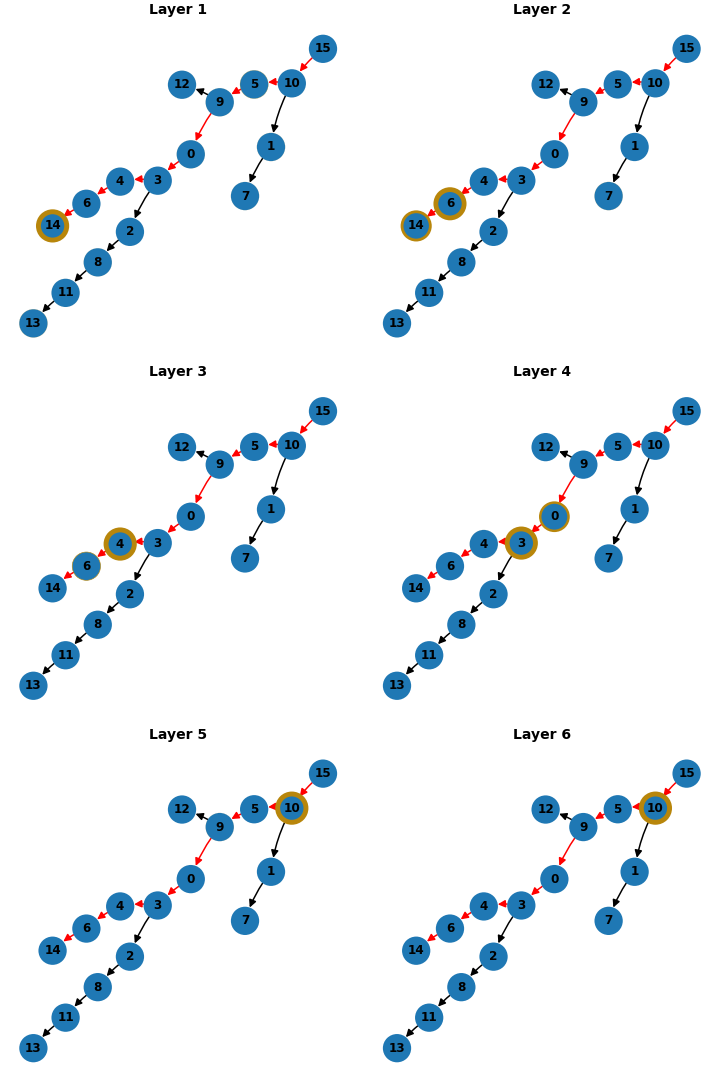} %
      \caption{Example 2 (Path Length 8): Results of a linear transformation to predict the next step based on the residual stream activations after each layer, projected onto the tree structure. The yellow border highlights the current best prediction(s) of the linear transformation.}
     \label{fig:tl2}
 \end{figure*}

\begin{figure*}[!h]
     \centering
     \includegraphics[width=0.9\textwidth]{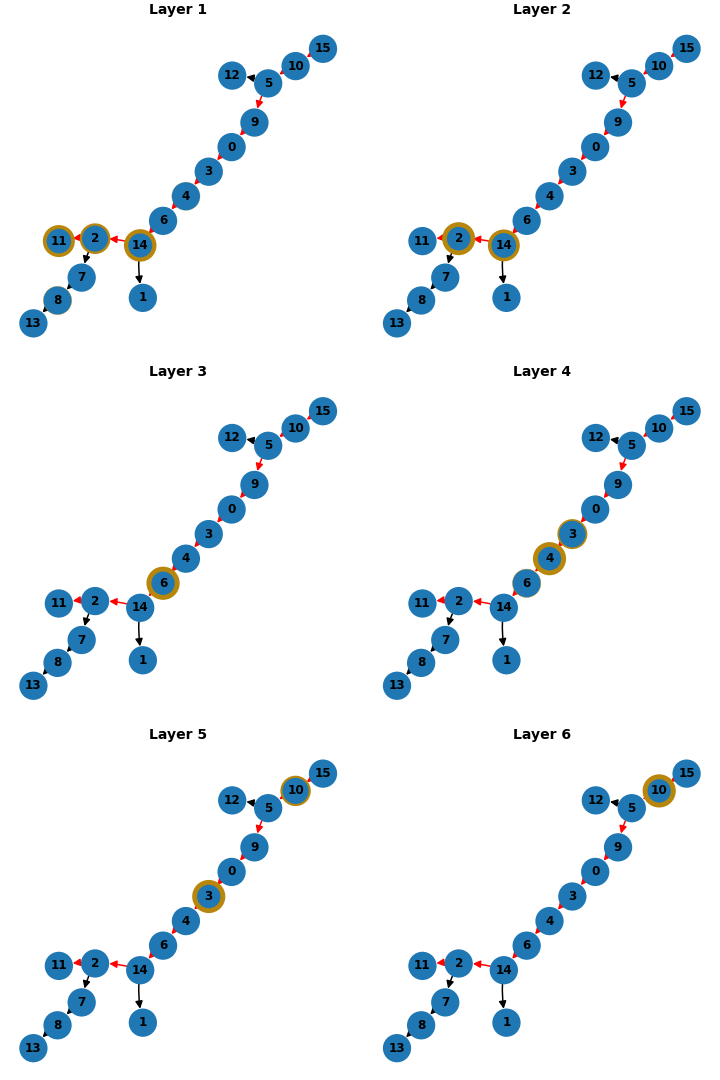} %
     \caption{Example 3 (Path Length 10): Results of a linear transformation to predict the next step based on the residual stream activations after each layer, projected onto the tree structure. The yellow border highlights the current best prediction(s) of the linear transformation.}
     \label{fig:tl3}
 \end{figure*}

\begin{figure*}[!h]
     \centering
     \includegraphics[width=0.9\textwidth]{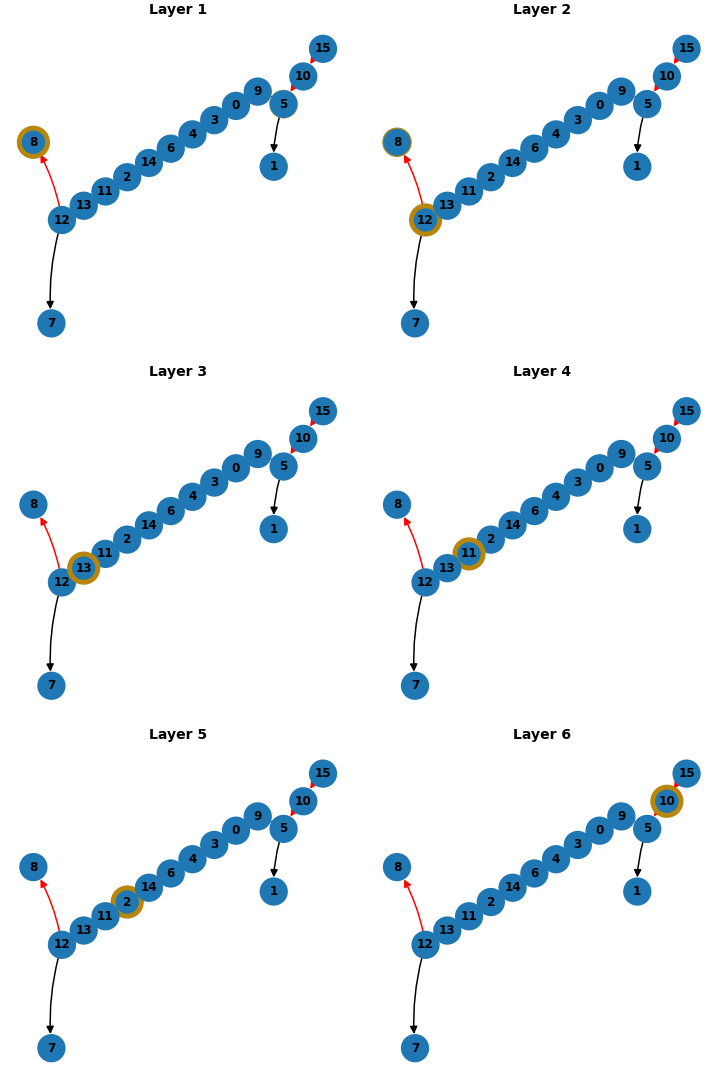} %
     \caption{Example 4 (Path Length 13): Results of a linear transformation to predict the next step based on the residual stream activations after each layer, projected onto the tree structure. The yellow border highlights the current best prediction(s) of the linear transformation.}
     \label{fig:tl4}
 \end{figure*}

Inspired by this approach, we replace the last $n$ layers of the model with a linear transformation trained to predict the next token from the residual stream activations $\mathbf{x}^{L-n}$.
Similar to the Tuned Lens, this method allows us to skip over these layers and see the current best prediction that can be made from the model's residual stream. 
Intuitively, this allows us to peek at the iterative computations a transformer uses to compute the next token. 
Here, we present a visualization of some example trees and the results of the iterative computation (see Figures~\ref{fig:tl1} to \ref{fig:tl4}). 
These figures highlight the current best prediction a linear transformation could make based on the internal activations.
We project the logits output by the linear model back onto the tree structure to better visualize the backward-chaining procedure.

\clearpage

\clearpage

\clearpage

\clearpage

\end{document}